\documentclass[10pt,twocolumn,letterpaper]{article}
\usepackage{iccv}
\usepackage{times}
\usepackage{epsfig}

\usepackage{graphicx}
\usepackage{amsmath}
\usepackage{amssymb}
\usepackage{epsfig}
\usepackage{graphicx}
\usepackage{booktabs}
\usepackage{textcomp}
\usepackage{algorithm}
\usepackage{booktabs}
\usepackage[noend]{algpseudocode}
\usepackage{tabularx}
\usepackage{tablefootnote}
\usepackage{listings}
\usepackage{multirow}
\usepackage{threeparttable}

\usepackage{xcolor}

\usepackage[accsupp]{axessibility}
\usepackage[pagebackref=true,breaklinks=true,letterpaper=true,colorlinks,bookmarks=false]{hyperref}

\usepackage[capitalize]{cleveref}
\crefname{section}{Sec.}{Secs.}
\Crefname{section}{Section}{Sections}
\Crefname{table}{Table}{Tables}
\crefname{table}{Tab.}{Tabs.}

\iccvfinalcopy 

\def\iccvPaperID{8707} 

\ificcvfinal\fi

\begin{document}

\title{Large-Scale Bidirectional Training for Zero-Shot Image Captioning}
\author{
Taehoon~Kim\thanks{\footnotesize Equal contribution.} $^\dagger$ \qquad Mark Marsden$^*$$^\ddagger$ \qquad Pyunghwan Ahn$^\dagger$ \qquad Sangyun Kim$^\dagger$ \\ \qquad Sihaeng Lee$^\dagger$ \qquad Alessandra Sala$^\ddagger$ \qquad Seung~Hwan~Kim$^\dagger$ \\
\\
LG AI Research\thanks{\footnotesize \{taehoon.kim, p.ahn, syun, sihaeng.lee, sh.kim\}@lgresearch.ai} \qquad \qquad Shutterstock\thanks{\footnotesize \{mmarsden, asala\}@shutterstock.com} \qquad 
}
\maketitle
\ificcvfinal\thispagestyle{empty}\fi
\begin{abstract}
When trained on large-scale datasets, image captioning models can understand the content of images from a general domain but often fail to generate accurate, detailed captions. To improve performance, pretraining-and-finetuning has been a key strategy for image captioning. However, we find that large-scale bidirectional training between image and text enables zero-shot image captioning. In this paper, we introduce Bidirectional Image Text Training in largER Scale, BITTERS, an efficient training and inference framework for zero-shot image captioning. We also propose a new evaluation benchmark which comprises of high quality datasets and an extensive set of metrics to properly evaluate zero-shot captioning accuracy and societal bias. We additionally provide an efficient finetuning approach for keyword extraction. We show that careful selection of large-scale training set and model architecture is the key to achieving zero-shot image captioning.
\end{abstract}

\section{Introduction}
\label{sec:intro}
Trained with billions of image-text pairs, large-scale vision-language models (LVLMs) are outperforming previous state-of-the-art approaches in visual understanding \cite{li2020oscar,zhang2021vinvl,hu2022lemon, wang2022git} and text-to-image generation \cite{ramesh2021zeroshot,ding2021cogview, ding2022cogview2, gafni2022makeasense,nichol2021glide, ramesh2022dalle2, yu2022parti, rombach2021stablediffusion}. Recent advancements in text-to-image generation models now allow us to generate high-quality, unseen images from a single text prompt. Unlike text-to-image generation  which focuses on zero-shot capability, image-to-text generation (image captioning) mainly relies on the pretraining-and-finetuning strategy. Even after pretraining on millions of image-text pairs \cite{hu2022lemon,wang2022git}, previous captioning models are observed to lack zero-shot capability as they struggle to generalize to a given domain without finetuning. On the other hand, unified image-text training methods \cite{kim2022verse, wang2022ofa} achieve notable results on various tasks. 

\begin{figure*}[!ht]
\vspace*{-0.05in}
  \centering
\includegraphics[width=\linewidth]{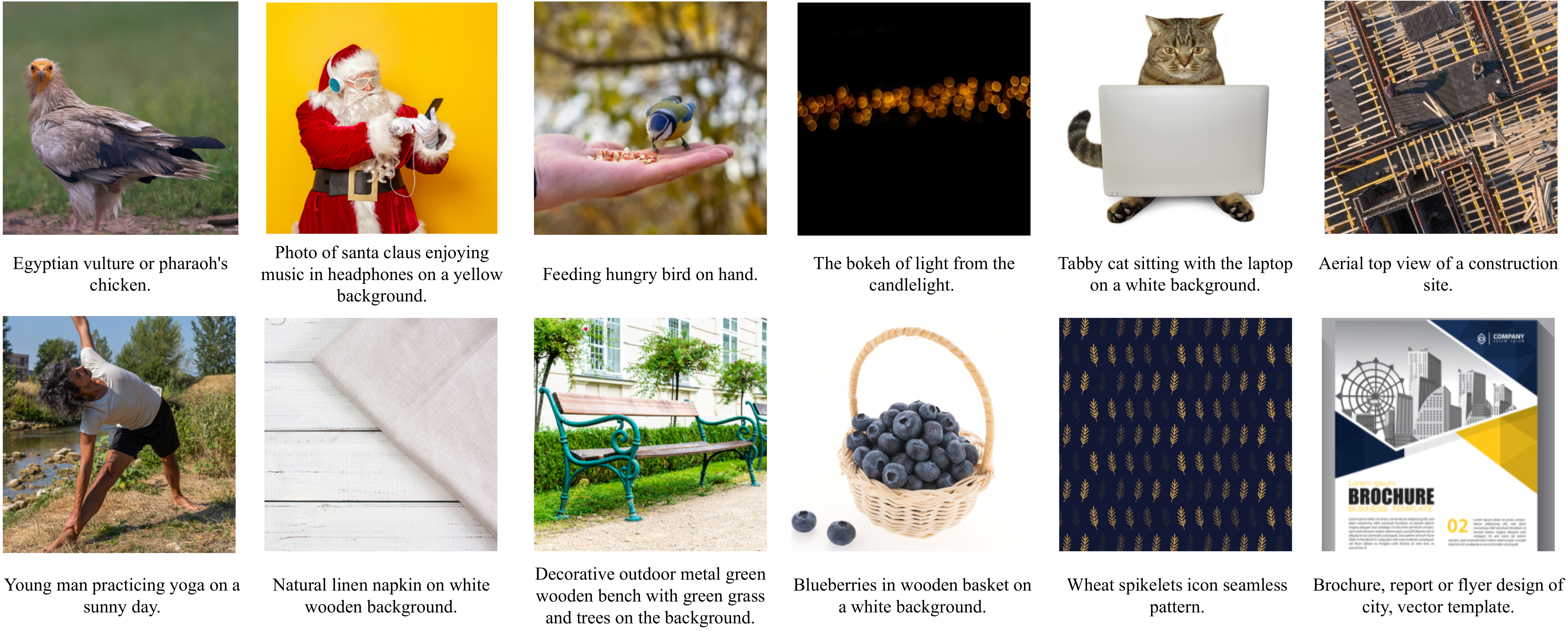}
\vspace*{-0.15in}
\caption{
Example captions generated with BITTERS. Our models shows zero-shot capability of image captioning by covering a wide variety of categories including photos, stock images, illustrations, vectors, as well as images of people from various demographic groups.}
\label{bitters}
\vspace{-4mm}
\end{figure*}

Inspired by the bidirectional training strategy introduced in \cite{kim2022verse}, we propose a large-scale training, inference, and evaluation framework for zero-shot image captioning. We find the main cause of poor zero-shot capability from the quality of training data. Texts in large-scale web-crawled datasets \cite{sharma2018conceptual,changpinyo2021cc12m, thomee2016yfcc,schuhmann2021laion400m,desai2021redcaps} have a wide variety of tones and manners. However, the \textit{scale} is also what makes it difficult to filter out slang, inappropriate languages, or poor descriptions. 
To gain control over generated captions, we train our model with a new collection of 100 million image-text pairs, specifically curated for zero-shot image captioning. As shown in Figure \ref{bitters}, our \textbf{B}idirectional \textbf{I}mage \textbf{T}ext \textbf{T}raining in larg\textbf{ER} \textbf{S}cale, \textbf{BITTERS}, generates detailed captions over a diverse set of image categories and styles.
Our contribution for the training and evaluation of a zero-shot image captioning model are as follows:
\begin{itemize}
    \item We propose a 2D discrete wavelet transform (DWT) based cross-level feature augmentation method with a new architecture for AugVAE \cite{kim2022verse}. Our improved AugVAE (WaveVAE) shows enhanced image reconstruction performance on the ImageNet1K \cite{imagenet_cvpr09} validation set. With WaveVAE and other changes to the BiART \cite{kim2022verse} parameter configuration, BITTERS shows 32\% less GPU memory usage and 18\% faster sampling speed compared to L-Verse \cite{kim2022verse}.
    
    \item We provide training (TIP100M) and evaluation (ICE-A and B) datasets for zero-shot image captioning. We also suggest various metrics to assess a given model's zero-shot image captioning performance and societal bias. We show how the distribution of the training set affects the quality and bias of generated captions.

    \item We further introduce an adapter-based \cite{houlsby2019adapter, hu2021lora, lin-etal-2020-exploring} finetuning approach for vision-language transformers. With mere 0.02\% increase in number of parameters and sampling speed to BITTERS, we also enable keyword extraction from images. 
    
\end{itemize}

\section{Related Work}
\label{sec:related_work}
In this section, we introduce large vision language models (LVLMs) for image-text generation tasks. Specifically, we introduce: \textit{(i)} VQ-VAE for image encoding and decoding, \textit{(ii)} LVLMs for image $\leftrightarrow$ text generation, and \textit{(iii)} a brief introduction on zero-shot image captioning. We also introduce more previous works related to BITTERS in our supplementary material.

\subsection{Vector Quantized Variational Autoencoder}
Vector quantized variational autoencoder, VQ-VAE \cite{oord2018neural}, is widely used to encode and decode images to and from a series of quantized vectors. VQ-VAEs consist of an encoder, a decoder, and a vector quantizer  with a visual codebook for learning discrete representations of images. As the vector quantizer factorizes the continuous representation of an image into a set of vectors within the limited possibilities of a visual codebook, information loss is inevitable. Reconstruction Fréchet Inception Distance (rFID) \cite{Heusel2017GANsTB, esser2021taming} is widely used to measure the difference between original and reconstructed images. 

From the assumption that the rFID of a VQ-VAE also effects the generation performance of a transformer it is attached to \cite{esser2021taming,ramesh2021zeroshot,yu2022vectorquantized, kim2022verse}, various methods have been proposed to improve rFID. Razavi \etal \cite{razavi2019generating} and Kim \etal \cite{kim2022verse} focus on hierarchical feature representations to update visual codebook with diverse features. Gumbel-softmax relaxation \cite{ramesh2021zeroshot} and $L_2$ normalization \cite{yu2022vectorquantized} are proposed to improve codebook usage. Combinations of $L_1$, $L_2$, logit-laplace \cite{ramesh2021zeroshot}, LPIPS \cite{zhang2018unreasonable}, or adversarial \cite{NIPS2014_5423} losses are proposed in \cite{ramesh2021zeroshot,kim2022verse,esser2021taming,yu2022vectorquantized} to directly reduce rFID.

\subsection{Image $\leftrightarrow$ Text Generation}
In the early stage of text-to-image generation, a combination of a VQ-VAE and an auto-regressive transformer were widely used \cite{ramesh2021zeroshot,ding2021cogview,gafni2022makeasense,yu2022parti} and showed promising results on zero-shot text-to-image generations. Recent works \cite{ramesh2022dalle2, nichol2021glide, ho2022imagen} further improve the generation quality by replacing the transformer with a denoising diffusion probabilistic model (DDPM) \cite{ho2020ddpm}. 

Unlike text-to-image, previous works \cite{li2020oscar, wang2022git, zhang2021vinvl, hu2022lemon, cho2021vlt5, mokady2021clipcap} on image-to-text generation (image captioning) focus on a pretraining-and-finetuning strategy. In accordance with LLMs, scaling up the pretraining data and the model size is important to build LVLMs. To control the generation result, the model is usually finetuned with a smaller dataset for each downstream-task. 

To learn a cross-modal relationship between image and text, unified image-text training approaches \cite{kim2022verse, wang2022ofa} have also been proposed. The bidirectional training strategy \cite{kim2022verse} shows data and parameter-efficient results compared to unidirectional models. OFA \cite{wang2022ofa} unifies modalities (vision, language, multimodal) and tasks for pretraining.

\subsection{Zero-Shot Image Captioning}
Tewel \etal \cite{tewel2022zerocap} proposes ZeroCap, a combination of pretrained multimodal (CLIP \cite{radford2021learning}) and language (GPT-2 \cite{radford2019language}) models to enable zero-shot image captioning. Unlike previous works mentioned above which require additional finetuning, ZeroCap performs inference time gradient updates and optimization of GPT-2 to generate a caption that matches the given image. This requires a CLIP text encoder forward per each word of the caption. Concurrently to this work, Li \etal \cite{li2023decap} decodes CLIP latents with a text-only trained language model for zero-shot image captioning.

\begin{figure*}[ht]

  \centering
\includegraphics[width=\linewidth]{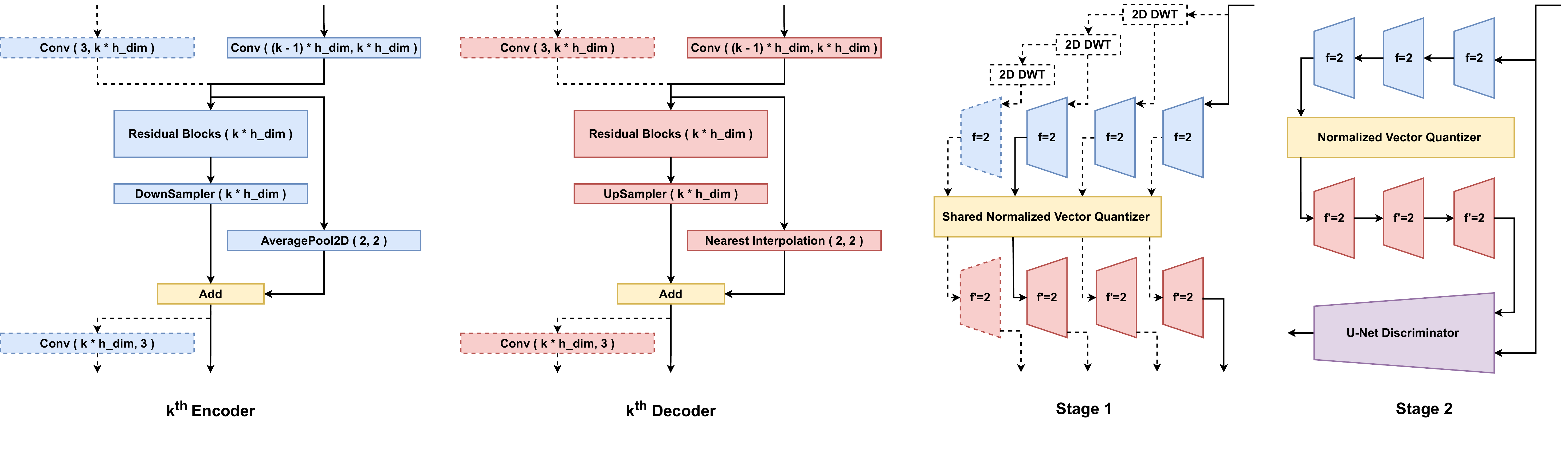}

\caption{
Proposed WaveVAE architecture. We remove unnecessary components (\textit{dotted line}) and run short calibration before Stage 2.}
\label{wavevae_figure}

\end{figure*}
\section{Method}
\label{sec:method}

\subsection{Preliminary}
Previous auto-regressive transformer \cite{Brown2020LanguageMA} based methods follow a two-stage training procedure proposed by Ramesh \etal \cite{ramesh2021zeroshot}  for image-to-text \cite{ kim2022verse, wang2022git} or text-to-image \cite{ramesh2021zeroshot, ding2021cogview, kim2022verse, ding2022cogview2, gafni2022makeasense} generation:

\begin{itemize}
\item \textbf{Stage 1:} Train a vector-quantized variational autoencoder (VQ-VAE) \cite{kingma2014autoencoding, razavi2019generating, ramesh2021zeroshot, esser2021taming, kim2022verse} to compress each RGB image into a sequence of image tokens with each element of \textit{$d_Z$} possible values.

\item \textbf{Stage 2:} Concatenate BPE-encoded text tokens and image tokens before training an auto-regressive transformer \cite{Brown2020LanguageMA} to model the joint distribution over text and image tokens. According to the order of text and image tokens, the transformer learns to generate an image from given text or vice-versa. 
\end{itemize}

For zero-shot image captioning, we bring the bidirectional image-text training concept of L-Verse \cite{kim2022verse} to a larger scale with 100 million image-text pairs. We also propose a new VQ-VAE, WaveVAE, which shows better image reconstruction quality with 75\% fewer parameters compared to AugVAE \cite{kim2022verse}.

\subsection{WaveVAE}
\label{wavevae}

According to Kim \etal \cite{kim2022verse}, similar patterns in various patch sizes can appear throughout the training. Kim \etal \cite{kim2022verse} proposes a two stage training theme for a VQ-VAE which utilizes this cross-level patch similarity: 
\begin{itemize}
\item \textbf{Stage 1:} Train a hierarchical AugVAE (AugVAE-ML) with cross-level feature augmentation. Four latent feature maps of different sizes are extracted to update a single vector quantizer.
\item \textbf{Stage 2:} Remove unnecessary components from AugVAE-ML and finetune the model into a single-level AugVAE (AugVAE-SL) of $32 \times 32$ latent map. 
\end{itemize}
 Following the notation from Kim \etal \cite{kim2022verse}, we define the $k^{th}$ encoder as $z = E_{k}(x, f,d_{in}, d_{out})$, where $x$ is an $n \times n \times d_{in}$ tensor, $f$ is a downsampling factor, and $z$ is an $\frac{n}{f} \times \frac{n}{f} \times d_{out}$ tensor. The vector quantizer is $z_q = VQ(z, d_{Z})$, where $z$ is an $n \times n \times d$ tensor with continuous $d$-size vectors and $z_q$ is a quantized version of $z$ with $d_{Z}$ possible values. The $k^{th}$ decoder is  $\hat{x} = G_{k}(\hat{z}, f', d_{in}', d_{out}')$, where $\hat{z}$ is an $n \times n \times d_{in}'$ tensor, $f'$ is an upsampling factor, and $\hat{x}$ is an $nf' \times nf' \times d_{out}'$ tensor. We further improve cross-level feature augmentation method in both theoretical and architectural aspects. 

While Kim \etal \cite{kim2022verse} directly use the output of $k^{th}$ encoder as the input for ${k+1}^{th}$ encoder for pretraining (Stage 1), we utilize 2D discrete wavelet transform (DWT) \cite{antonini1992dwt} to generate an input for each encoder. The 2D DWT decomposes an image into three high-pass filtered images, each describing local changes in details, and one low-pass filtered image. The low-pass filtered image is a downscaled approximation of the original image which can replace the output of previous encoder. Specifically, our DWT applied AugVAE (WaveVAE) is trained as follows:
\begin{itemize}
\item \textbf{Stage 1:} Pair each $E_{k}(2, 3, k * hidden\_dim)$ with $G_{k}(2, k * hidden\_dim, 3)$. For each pair, apply DWT decomposition $k-1$ times to generate input $x_{k}$. All pairs of encoder and decoder shares one $VQ(8192)$ for cross-level feature augmentation.
\item \textbf{Stage 2:} Integrate encoders into a single encoder $E(8, 3, 3 * hidden\_dim)$ and decoders into a single decoder $G(8,3* hidden\_dim, 3)$. After removing redundant components, calibrate the single-level encoder - vector quantizer - decoder architecture for few iterations and finetune the model with various loss terms.
\end{itemize}

Along with DWT-based cross-level feature augmentation, we optimized the architecture of proposed WaveVAE for better image reconstruction quality with fewer parameters. We reduced the number of parameters to 25 million in total, a 75\% reduction compared to AugVAE-SL. 
\begin{figure*}[!tb]
  \centering
\includegraphics[width=\linewidth]{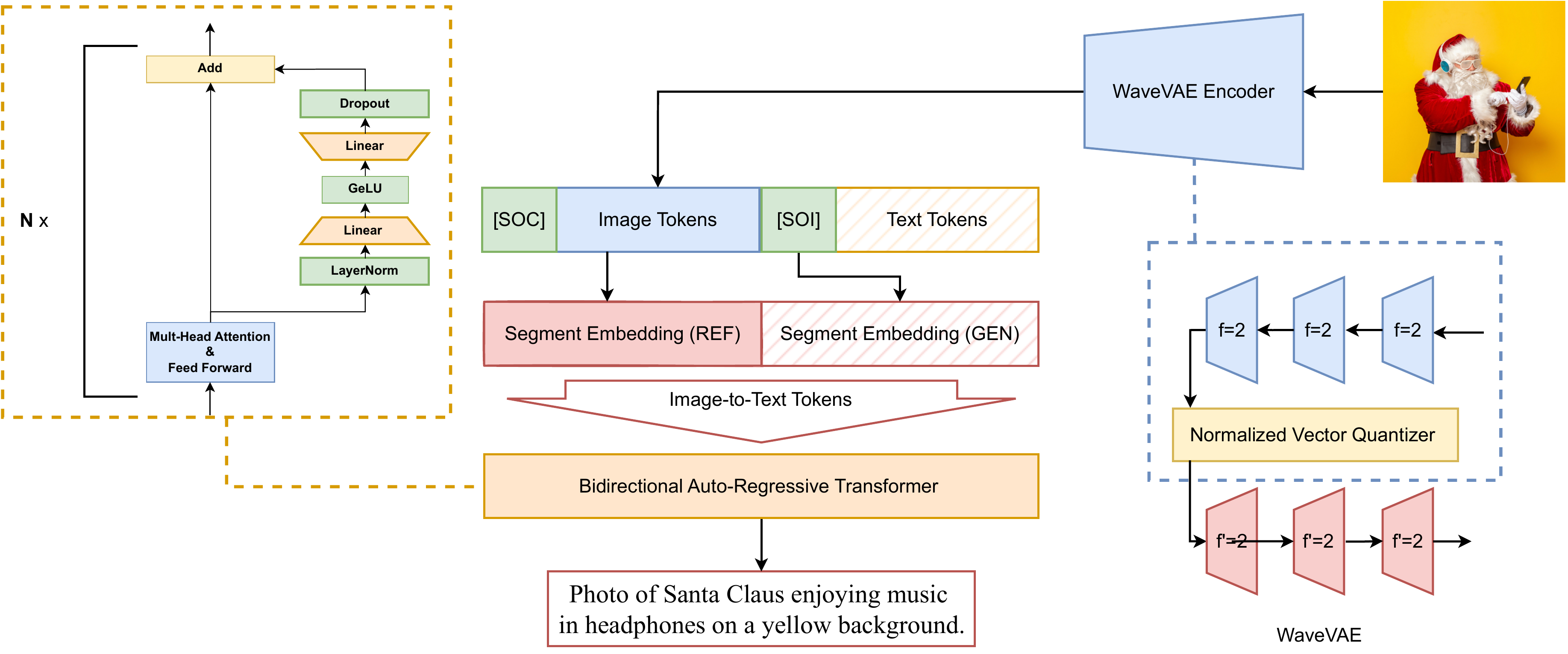}
\caption{Proposed BITTERS architecture with the adapter for finetuning. Adapter is attached to each layer of Bidirectional Auto-Regressive Transformer (BiART) for parameter-efficient finetuning.
}
\label{adapter}

\end{figure*}

\paragraph{Architecture}
The architecture of the proposed WaveVAE is depicted in Figure \ref{wavevae_figure}. The encoder and decoder of WaveVAE are both bottleneck-style residual network similar to AugVAE \cite{kim2022verse}. Unlike AugVAE, we use Parametric Rectified Linear Unit (PReLU) \cite{he2015prelu} as the activation function. We also use PixelShuffle \cite{shi2016pixelshuffle} for upsampling. We use encoders and decoders with 8 residual blocks and the hidden dimension (\textit{hidden\_dim}) of 64. For the vector quantizer,  we use a visual codebook with the embedding dimension of 64 and the codebook size of 8192. We do not use an exponential moving average (EMA) vector quantizer \cite{razavi2019generating} as in \cite{kim2022verse}. Instead, we use an $L_2$ normalized vector quantizer as proposed in \cite{yu2022vectorquantized}. Precise details including the source code are provided in our supplementary material.

\subsection{BiART}
\label{training_detail}

\paragraph{Architecture} Along with WaveVAE, we also modify the parameter setting of BiART \cite{kim2022verse} according to the scaling law proposed in \cite{hoffmann2022chinchilla}. BITTERS uses a 24-layer BiART \cite{kim2022verse} with 1280 dimensional states and 10 masked self-attention heads. We use 64 BPE-encoded \cite{sennrich2016neural} text tokens with 49408 possibilities and 1024 encoded image tokens with 8192 possibilities. BITTERS has 650 million parameters in total. Compared to L-Verse \cite{kim2022verse}, BITTERS shows 32\% less memory usage and 18\% faster inference speed on a single NVIDIA A100 GPU. While BiART is initially designed for bidirectional training between image and text, we adapt the bidirectional training concept of BiART \textit{only for stable training}. As stated in Kim \etal \cite{kim2022verse}, bidirectional training relieves the heterogeneity between image and text, and prevents overflow of gradients. We also tried to train a model only in image-to-text direction but the model diverged.

\paragraph{Parameter-Efficient Finetuning} 
Unlike previous methods \cite{li2020oscar, wang2022git, zhang2021vinvl,hu2022lemon, cho2021vlt5}, we don't update all weights during finetuning as this risks modifying the learned joint distribution between image and text. Among the various finetuning approaches \cite{liu2021ptuning, liu2021ptuningv2, houlsby2019adapter, hu2021lora, lin-etal-2020-exploring} for transformers, we take an adapter-based approach \cite{houlsby2019adapter, hu2021lora, lin-etal-2020-exploring}. As shown in Figure \ref{adapter}, we attach an adapter with bottleneck dimension of 320 to each layer of BiART. During finetuning, we only update adapters with hyperparameters used for pretraining. Using adapters brings only 0.02\% increase in number of parameters and latency.

\subsection{Sampling}
\label{sampling}
\paragraph{Image Captioning} As we focus on zero-shot image captioning, we modify the text sampling process proposed in \cite{kim2022verse} to delicately control the generated caption. We sample 32 text tokens with pretrained BITTERS model to generate a caption for each image. For each token selection, we first select 10\% of logits with the highest probabilities (\textit{top-k} sampling) \cite{fan2018topk} and apply \textit{top-p} sampling \cite{Holtzman2020topp} with $p=0.95$.  We sample 64 captions in total and calculate CLIPScore \cite{hessel-etal-2021-clipscore} to select a Top-1 caption. 

\paragraph{Keyword Extraction} To extract keywords from image, we sample 48 text tokens with finetuned BITTERS model to generate a list of keywords for each image. We sample 64 lists of keywords in total and calculate CLIPScore \cite{hessel-etal-2021-clipscore} to select a Top-8 lists of keywords. From Top-8 lists of keywords, we further select keywords that appeared 3 or more times in the Top-8 lists. Other details are identical to the sampling method for image captioning.

\section{Dataset} 
\label{sec:dataset}
In this section, we briefly introduce our newly curated training and evaluation sets for zero-shot image captioning. More details including license and level of publicity for each dataset are provided in our supplementary material.  

\subsection{Training}
\label{sec:dataset_tip}
As stated in \cite{kim2022verse, wang2022git}, it is difficult to control the generated caption from a model trained with a large number of web-harvested image-text pairs. This is due to the wide variety of text styles, inappropriate language, and intrinsic biases present. For this reason, we train BITTERS using a new, quality-controlled 100 million image dataset which we will refer to as Text Image Pairs 100 Million (TIP100M). 

All images are random sampled from Shutterstock's\footnote{\href {https://www.shutterstock.com}{www.shutterstock.com}} image catalog. Images are 500px on the longest side with a single ground truth caption provided. A list of keywords is also included for each image to enable keyword extraction model training. All captions and keywords are in English and were moderated for hate speech, slurs, and expletives. 

This dataset contains a very broad set of concepts and scenarios (indoor and outdoor, with or without people, with or without animals). The majority of images are photographs (69\%) while the rest are either illustrations or vector graphics (31\% combined). Roughly 1/3 (24/69\%) of included photographs contain at least one person. For photographs containing people: 75\% are Caucasian, 53\% are 20-30 years of age and 67\% are female. 10\% of images are editorial use content (containing logos, celebrities, news content). 

\subsection{Evaluation} 
\label{sec:dataset_ice}
Evaluation is carried out using two newly curated image datasets. These collections are produced to better assess the true limits of zero-shot image captioning models by identifying where a given model succeeds or fails to produce accurate and fair image captions. 
We refer to each dataset as Image Captioning Evaluation Accuracy (ICE-A) and Image Captioning Evaluation Bias (ICE-B) respectively. 

\paragraph{ICE-A} This collection consists of 17 image categories from 4 high-level groups (People, Animals, Objects, Other). There is a varied set of 300 curated images per category, 5100 images in total.  To avoid redundancy, the full set of category names are outlined in Table \ref{table:dataset_a_cat_spice}. All image categories consist of photos unless otherwise stated (i.e. vector graphics, illustrations). In this context, `Stocky Setting' refers to photographs taken in a very controlled setting such as a studio. In comparison,`Authentic' refers to photographs taken in a more natural, candid setting, typically with natural lighting.

\paragraph{ICE-B} This collection was generated to expand upon the `Various Demographics' category in ICE-A, allowing for an in depth assessment of societal bias in a manner similar to \cite{hirota2022quantifying}. During the creation of this dataset, the following constraints were enforced: \textit{(i)} each image must be a photograph containing exactly one person and \textit{(ii)} demographic metadata (gender, ethnicity) must be available for each image. Enforcing both constraints allows us to directly assess  societal biases and split the dataset into demographic subgroups. 
The dataset consists of 14K female subject images and 8K of male subjects\footnote{We use a binary simplification of gender. We acknowledge that this is not inclusive or representative and should be addressed in future work.}. The following ethnicity labels breakdown is observed:  Southeast Asian (5686), Caucasian (4782), Hispanic (3965), African American (3878), East Asian (2136), South Asian (1240), Middle Eastern (704)\footnote{We use a fixed set of 7 race/ethnicity groups. We acknowledge that this set of groups is far from comprehensive and that these attributes are much more complex in reality.}. Beyond diversity of human subjects, this dataset also contains a variety of scenarios and concepts including indoor, outdoor, facial masks being worn, hand gestures, medical settings, exercising, workplace, and family home.

\section{Metrics}
\label{sec:metric}
In this section, we briefly introduce quantitative measures that we use to assess the zero-shot image captioning performance of our model. More details on each evaluation metric are provided in our supplementary material. 

\paragraph{Caption Accuracy} We evaluate caption accuracy using a set of commonly employed metrics: SPICE \cite{anderson2016spice}, CIDEr \cite{vedantam2015cider}, METEOR \cite{banerjee2005meteor}, ROGUE \cite{lin2004rouge} and BLEU-4 \cite{papineni2002bleu}. For qualitative assessment, we also conduct a human evaluation on BITTERS generated captions along with generations from from other models. 

\paragraph{Bias Assessment} To prevent potential negative impacts of zero-shot image captioning, we also put emphasis on detecting gender or racial prejudice in generated captions. We assess the societal bias of a given model using three distinct approaches: Gender Error and Term Ratio \cite{hendricks2018women}, VADER Sentiment Score \cite{zhao2021understanding}, and Leakage for Image Captioning (LIC) \cite{hirota2022quantifying}. 

\paragraph{Keyword Extraction} We evaluate the keyword extraction performance of BITTERS on image-keywords ground truth pairs from ICE-A using the following two metrics: Normalized Keyword Overlap and CLIP Cosine Similarity. To show the relevance between image captioning and keyword extraction, we use the same set of example images for generated captions and extracted keywords in our supplementary material.

\begin{figure*}[tb]
  \centering
\includegraphics[width=0.98\linewidth]{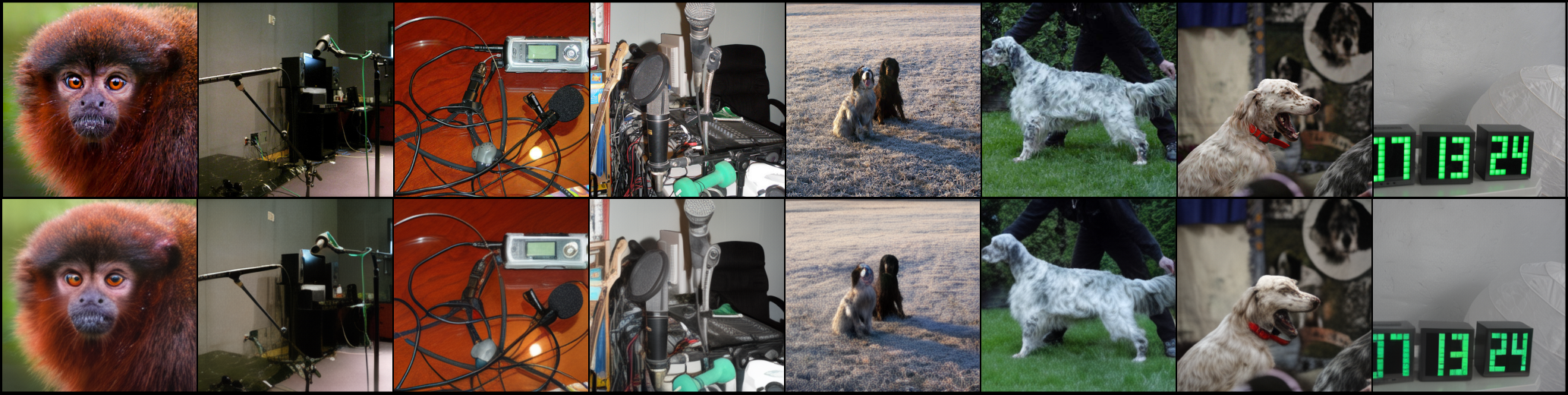}
\caption{
Examples of input images \textit{(top)}
and reconstructions from WaveVAE \textit{(bottom)}. The resolution of each
image is $256 \times 256$.}

\end{figure*}
\section{Experiment}
\label{sec:experiments} 

\subsection{Image Reconstruction}
\label{sec:exp_recon}
From the result in Table \ref{table:recon_fid}, WaveVAE shows better (lower) reconstruction Fréchet Inception Distance \cite{Heusel2017GANsTB} (rFID) on the ImageNet1K \cite{imagenet_cvpr09} validation set when trained in every case with different training datasets. It is also notable that WaveVAE shows improvement in rFID when trained with a larger dataset (TIP100M), while AugVAE-SL shows performance degradation. This suggests that WaveVAE is more suitable than AugVAE-SL for large-scale training.

\begin{table}[t]
\centering
\footnotesize
\addtolength{\tabcolsep}{-2pt}
\begin{tabular}{lcccc}
\toprule
Model              & Training Data  & rFID & \\ 
\midrule
AugVAE-SL \cite{kim2022verse}             & ImageNet1K  &  3.28   &\\
WaveVAE             & ImageNet1K    & \textbf{2.70}   &\\
\midrule
AugVAE-SL \cite{kim2022verse}             & TIP100M   & 3.94   &\\
WaveVAE             & TIP100M    & \textbf{2.35}   &\\
\midrule

\end{tabular}
\caption
{
Reconstruction Fréchet Inception Distance (rFID) on ImageNet1K validation set. Our WaveVAE shows better (lower) rFID in all cases compared to AugVAE-SL with 75\% less parameters.
}
\label{table:recon_fid}
\end{table}

\subsection{Zero-Shot Image Captioning} 
\label{image_text}
\paragraph{Result on Public Benchmark}
 We first compare BITTERS with ZeroCap \cite{tewel2022zerocap}, current state-of-the-art in zero-shot image captioning, on MS-COCO Captions \cite{lin2015microsoft} \textit{karpathy test split} and ICE-A. From the results in Table \ref{table:rebuttal}, BITTERS outperforms ZeroCap on MS-COCO Captions in all metrics. Due to gradient update and optimization over the context cache, ZeroCap shows 30 times slower inference speed compared to BITTERS. From the results in Tables \ref{table:rebuttal} and \ref{table:dataset_a_perf_overall}, ZeroCap shows poor result on ICE-A even compared to ClipCap \cite{mokady2021clipcap} and GIT \cite{wang2022git}. ZeroCap fails to generate proper captions for images in categories such as vectors, stylized, illustrations, and medical equipment, \textit{which is not included in MS-COCO}. This demonstrates the importance of a new evaluation set with diverse image categories (ICE-A) to assess a given model’s zero-shot image captioning performance.

\paragraph{Experimental Setup} We perform quantitative evaluation on four model architectures. Two model architectures trained on TIP100M, L-Verse \cite{kim2022verse} and the proposed BITTERS, are set as an experimental group. We also include ClipCap \cite{mokady2021clipcap} and GIT (\texttt{BASE\_MODEL}) \cite{wang2022git} as a control group. These models are chosen for comparison as they are trained on large-scale datasets and perform well on other captioning benchmarks.
 \begin{table}[t]
\centering
\footnotesize
\addtolength{\tabcolsep}{-2pt}
\begin{threeparttable}
\begin{tabular}{lccccc}
 & \multicolumn{2}{c}{MS-COCO} & \multicolumn{2}{c}{ICE-A} \\
\toprule
Metric       & ZeroCap \cite{tewel2022zerocap}   & BITTERS  & ZeroCap \cite{tewel2022zerocap}     & BITTERS   & \\ 
\midrule
BLEU-4       &  2.6    &  \textbf{6.1}  & 0.24  &  \textbf{2.9} & \\
METEOR       &  11.5   &  \textbf{13.7}    & 3.5   &\textbf{11.7}  & \\
ROUGE        &   -     &   27.8  &  8.1  &  \textbf{18.3} &\\
CIDEr        &  14.6   &  \textbf{31.2}  & 5.7   & \textbf{35.3}  &\\
SPICE        &  5.5    &  \textbf{9.5}  & 4.8   & \textbf{13.2} &\\
\midrule
\end{tabular}
\end{threeparttable}
\caption
{
Zero-shot image captioning accuracy of ZeroCap and BITTERS on MS-COCO Captions \textit{karpathy test split} and ICE-A.
}
\vspace{-4mm}
\label{table:rebuttal}
\end{table}

\paragraph{Caption Accuracy}
\label{caption_acc}
Table \ref{table:dataset_a_perf_overall} shows the overall caption accuracy metrics for ICE-A. L-Verse performs best in terms of BLEU-4, CIDEr and SPICE. We also use SPICE score to compare caption accuracy over different categories. In Table \ref{table:dataset_a_cat_spice}, L-Verse performs best overall (highest SPICE score in 9/17 categories) with BITTERS a close second. Regarding computational efficiency as mentioned in Section \ref{training_detail}, BITTERS shows comparable performance to L-Verse in all metrics. We find that the distribution of contents in a training set highly influences generated captions. While GIT performs noticeably better on `Food', ClipCap performs best on several categories in `Animals' group. On the other hand, L-Verse and BITTERS lead on `Vector Graphics', `Illustrations', and `Stylized' by a large margin. Some content is likely out of domain for ClipCap and GIT (i.e. `Stylized') hence the large difference in SPICE score.

For ICE-B, SPICE score per gender group is in Table \ref{table:dataset_b_gender_perf} and SPICE score per ethnicity group is in Table \ref{table:dataset_b_ethnicity_perf}. L-Verse and BITTERS again outperform the approaches in the control group. While the internal performance gap between gender groups is minimal for all models, there is a much larger variation between ethnicity groups. It is also observed that models which perform better in terms of overall SPICE score (L-Verse, BITTERS) have much larger variation in SPICE score between ethnicity groups.
\begin{table}[t]
\centering
\footnotesize
\addtolength{\tabcolsep}{-2pt}
\begin{threeparttable}
\begin{tabular}{lccccc}
\toprule
Metric             & ClipCap \cite{mokady2021clipcap}    & GIT \cite{wang2022git}   & L-Verse \cite{kim2022verse}     & BITTERS   & \\ 
\midrule
BLEU-4       &  0.8   &  0.7   & \textbf{3.6}   &  2.9 & \\
METEOR        & 7.2  & 5.4  & \textbf{11.7} &\textbf{11.7}  & \\
\midrule
ROUGE         & 14.2  & 12.1 &  18.1 &  \textbf{18.3} &\\
CIDEr          &  21.4  &  19.9 & \textbf{37.8} & 35.3  &\\
SPICE          &   11.9 &  10.5 & \textbf{13.8} & 13.2 &\\
\midrule
\end{tabular}
\end{threeparttable}
\caption
{
Zero-shot image captioning accuracy on ICE-A.
}
\label{table:dataset_a_perf_overall}
\end{table}
\begin{table}[t]
\centering
\footnotesize
\addtolength{\tabcolsep}{-2pt}
\begin{threeparttable}[t]
\begin{tabular}{lccccc}
\toprule
Category & ClipCap \cite{mokady2021clipcap}  & GIT \cite{wang2022git} & L-Verse \cite{kim2022verse}  & BITTERS  \\
\midrule
P: AuI  & \textbf{13.6} & 7.6  & 13.2  & 12.3  \\
P: AuO &  \textbf{13.4} & 13.5  & 11.2  & 13.1  \\
P: StS & 11.9  & 13.6  & 16.3  & \textbf{20.1}  \\
P: VaD & 12.7  & 9.1  & \textbf{16.5}  & 13.1  \\
\midrule
A: AuI & \textbf{17.2}  &  10.3 & 10.8 & 14.6 \\
A: AuO  & 9.9  & 10.9  & \textbf{12.2}  & 11.5  \\
A: StS & \textbf{ 22.2}  & 17.4  & 21.2  & 21.3  \\
Ob: CoS & 10.5 & 8.2  & \textbf{13.2} & 12.3 \\
Ob: CoU & 14.1  & 13.2  & \textbf{16.3}  & 14.9  \\
Ob: Ele & 10.7  & 10.0  & \textbf{14.4} & 11.8  \\
Ob: Food & 13.7 & \textbf{20.2}  &  18.8 & 15.3  \\
Ob: Fur & 12.5  & 9.4  & 10.9  & \textbf{13.2}  \\
Ob: MeE  & \textbf{8.6 } & 4.9  & 8.2  & 7.8  \\
Ob: RdS & \textbf{10.4}  & 9.1  & 8.9 & 8.3  \\
\midrule
Ot: Ill  & 8.2  & 8.6  & \textbf{15.1}  & 10.9  \\
Ot: Sty & 5.8  & 5.5  & \textbf{17.8}  & 16.8  \\
Ot: VeG & 7.0  & 6.5  & \textbf{12.1}  & 10.3  \\
\midrule
\end{tabular}
\begin{tablenotes}[flushleft]
   \item[*] \footnotesize{\textbf{P}: People \textbf{A}: Animals \textbf{Ob}: Objects \textbf{Ot}: Other }
   \item[**] \footnotesize{\textbf{AuI}: Authentic Indoors \textbf{AuO}: Authentic Outdoors \textbf{StS}: Stocky Setting \textbf{VaD}: Various Demographics \textbf{CoS}: Construction Site \textbf{CoU}: Cooking Utensils \textbf{Ele}: Electronics \textbf{Fur}: Furniture \textbf{MeE}: Medical Equipment \textbf{RdS}: Road Signs \textbf{Ill}: Illustrations \textbf{Sty}: Stylized \textbf{VeG}: Vector Graphics}
\end{tablenotes}

\end{threeparttable}
\vspace*{1mm}
\caption
{
SPICE score per image category on ICE-A.
}
\label{table:dataset_a_cat_spice}
\end{table}

\paragraph{Human Evaluation}
We further conduct a human evaluation over generated captions on ICE-A. We use a web-based human evaluation tool as shown in our supplementary material. 100 anonymous evaluators participated in this research. For each participant, we provided 20 images randomly sampled from ICE-A with a caption generated from each of the four models (ClipCap, GIT, L-Verse, BITTERS). We asked them to choose the most appropriate caption for each image. We also allowed each participant to choose \textit{``None of the captions well describes the image."} Evaluation results are provided in Table \ref{table:human_eval}. 
With the question \textit{``Which caption best describes the given image?"} asked, BITTERS generated captions received \textbf{29.9\%} of the votes among other choices including \textit{``None of the captions well describes the image."} (14.3\%). BITTERS generated captions also had higher preference (\textbf{34.9\%}) by a large margin compared to the other three models. 

\begin{table}[t]
\centering
\footnotesize
\addtolength{\tabcolsep}{-2pt}
\begin{tabular}{lccccc}
\toprule
Group & ClipCap \cite{mokady2021clipcap}  & GIT \cite{wang2022git} & L-Verse \cite{kim2022verse}  & BITTERS \\
\midrule
Male & 12.2 & 9.8  & \textbf{16.3 } & 15.4  \\
Female & 12.5  & 10.4  & \textbf{16.9 } & 16.1  \\
All &  12.3 & 10.2  & \textbf{16.7}  & 15.9 \\
\midrule
\end{tabular}

\caption
{
SPICE score per gender group on ICE-B.
}
\label{table:dataset_b_gender_perf}
\end{table}
\begin{table}[t]
\centering
\footnotesize
\addtolength{\tabcolsep}{-2pt}
\begin{threeparttable}[t]
\begin{tabular}{lccccc}
\toprule
Group & ClipCap \cite{mokady2021clipcap}  & GIT \cite{wang2022git} & L-Verse \cite{kim2022verse}  & BITTERS\\
\midrule
AF-AM & 11.6 &   9.4 & \textbf{18.6}  & 17.2 \\
C & 12.5 & 10.5 & \textbf{14.6} & 13.9 \\
E-A &  12.1 & 9.6  & 12.7  & \textbf{13.0} \\
H &  11.7 & 10.9   & \textbf{22.8}  & 20.8 \\
M-E & 11.3  & 9.2  & 13.5  & \textbf{13.5} \\
S-A & 17.1  & 13.4   & \textbf{27.9}  & 24.4 \\
SE-A & 12.4  & 9.7  & 12.4  & \textbf{12.9 }\\
All &  12.3 & 10.2  & \textbf{16.7}  & 15.9 \\
\midrule
\end{tabular}
\begin{tablenotes}
   \item[*] \footnotesize{\textbf{AF-AM}: African American \textbf{C}: Caucasian  \textbf{E-A}: East Asian \textbf{H}: Hispanic  \textbf{M-E}: Middle Eastern \textbf{S-A}: South Asian  \textbf{SE-A}: Southeast Asian}
\end{tablenotes}

\end{threeparttable}
\vspace*{1mm}
\caption
{
SPICE score per ethnicity group on ICE-B.
}
\label{table:dataset_b_ethnicity_perf}
\end{table}
\begin{table}[t]
\centering
\footnotesize
\addtolength{\tabcolsep}{-2pt}
\begin{threeparttable}[t]
\begin{tabular}{lccccc}
\toprule
Percentage (\%) & ClipCap \cite{mokady2021clipcap}  & GIT \cite{wang2022git} & L-Verse \cite{kim2022verse}  & BITTERS \\
\midrule
 w/ None & 18.2  & 16.9  & 20.7   & \textbf{29.9}   \\
w/o None &  21.2 & 19.7   & 24.2  & \textbf{34.9}  \\
\midrule
\end{tabular}
\begin{tablenotes}[flushleft]
   \item[*] \footnotesize{\textbf{w/ None}: Percentage including votes for \textit{``None of the captions well describes the image"} (\textit{None}, 14.3\%) \textbf{w/o None}: Percentage without votes for \textit{None}. }

\end{tablenotes}
\end{threeparttable}
\vspace*{1mm}
\caption
{
Human evaluation results on ICE-A. 100 anonymous participants are asked the question \textit{``Which caption best describes the given image?"} for 20 random images sampled from ICE-A. 
}
\label{table:human_eval}
\end{table}

\paragraph{Bias Assessment}

Table \ref{table:dataset_b_bias_gender_error} presents gender error and ratio results on ICE-B. GIT has the lowest gender error rate while the proposed BITTERS has the highest. All four models have a gender term ratio similar to the true ratio of female to male subjects (1.75). While GIT and ClipCap generated captions are less prone to gender error,  they are also observed to be shorter and less semantically accurate (i.e. SPICE score). There appears to be an accuracy-bias trade-off for image captioning models. As shown in \cite{hirota2022quantifying}, other image captioning models also have similar gender error rates of 2-4\%. Overall results suggest that gender bias is a common issue in image captioning.

Neutral sentiment rate (\%) per ethnicity group and gender group are shown in Table \ref{table:dataset_b_ethnicity_sentiment} and Table \ref{table:dataset_b_gender_sentiment}. Models trained on TIP100M (L-Verse, BITTERS) contain more sentimental terms in their captions compared to ClipCap and GIT, which both show a much higher neutral sentiment rate. All models tend to generate more sentimentally neutral captions for images of male subjects, while they use more emotional languages for female subjects. Images of Hispanic subjects have the lowest neutral sentiment rate for all captioning models. The use of sentimental terms (i.e. `sad',`angry',`happy') in image captions is not inherently good or bad as this language is often required to properly describe the image contents. However, large differences between gender or ethnicity groups are undesirable (i.e. showing strong tendency to use emotive language to describe certain gender or ethnicity of a person being photographed).
\begin{table}[t]
\centering
\footnotesize
\addtolength{\tabcolsep}{-2pt}
\begin{threeparttable}[t]
\begin{tabular}{lccccc}
\toprule
Metric             & ClipCap \cite{mokady2021clipcap}    & GIT \cite{wang2022git}   & L-Verse \cite{kim2022verse}     & BITTERS   & \\
\midrule
Error (\%) $\downarrow$       & 3.1   & \textbf{1.8}   & 4.7 & 4.8    \\
Term Ratio           & 2.01   & 1.83  & 1.82 & 1.88  \\
\midrule
\end{tabular}

\end{threeparttable}
\caption
{
Gender error (\%) and term ratio on ICE-B. Gender term ratio for ground truth captions is 1.75.
}
\label{table:dataset_b_bias_gender_error}
\end{table}
\begin{table}[t]
\centering
\footnotesize
\addtolength{\tabcolsep}{-2pt}
\begin{threeparttable}[t]
\begin{tabular}{lccccc}
\toprule
Group & ClipCap \cite{hirota2022quantifying}  & GIT \cite{wang2022git} & L-Verse \cite{kim2022verse}  & BITTERS \\
\midrule
AF-AM &  67.6 & 89.7  & 15.2   & 14.8  \\
C &  75.5 & 91.4  & 29.3 & 26.5 \\
E-A & 82.1  & 91.8  & 27.9   &  28.3 \\
H &  59.7  & 85.8    & 10.7  & 10.9  \\
M-E & 75.8  & 95.5   & 21.4   &  22.4 \\
S-A & 69.3   & 95.5   & 20.2   & 20.9 \\
SE-A & 83.8   & 91.5   & 32.4   & 29.9\\
\midrule
\end{tabular}
\begin{tablenotes}
   \item[*] \footnotesize{\textbf{AF-AM}: African American \textbf{C}: Caucasian  \textbf{E-A}: East Asian \textbf{H}: Hispanic  \textbf{M-E}: Middle Eastern \textbf{S-A}: South Asian  \textbf{SE-A}: Southeast Asian}
\end{tablenotes}

\end{threeparttable}
\vspace*{1mm}
\caption
{
Neutral sentiment rate (\%) per ethnicity group on ICE-B.
}
\label{table:dataset_b_ethnicity_sentiment}
\end{table}
\begin{table}[t]
\centering
\footnotesize
\addtolength{\tabcolsep}{-2pt}
\begin{threeparttable}[t]
\begin{tabular}{lccccc}
\toprule
Group & ClipCap \cite{mokady2021clipcap}  & GIT \cite{wang2022git} & L-Verse \cite{kim2022verse}  & BITTERS \\
\midrule
Male &  77.6  & 91.0  & 26.4   & 25.7   \\
Female & 71.5  & 90.2   & 21.8  & 20.3  \\
\midrule
\end{tabular}

\end{threeparttable}
\caption
{
Neutral sentiment rate (\%) per gender group on ICE-B.
}
\label{table:dataset_b_gender_sentiment}
\end{table}
\begin{table}[t]
\centering
\footnotesize
\addtolength{\tabcolsep}{-2pt}
\begin{threeparttable}[t]
\begin{tabular}{lccccc}
\toprule
Metric             & ClipCap \cite{mokady2021clipcap}    & GIT \cite{wang2022git}   & L-Verse \cite{kim2022verse}     & BITTERS   & \\
\midrule
LIC    &      \textbf{44.9} & 45.3 & 49.9 & 52.0     \\
\midrule
\end{tabular}

\end{threeparttable}
\caption
{
Gender bias LIC scores on ICE-B. LIC score for ground truth captions is 49.2.
}
\label{table:dataset_b_bias_LIC}
\end{table}

Table \ref{table:dataset_b_bias_LIC} presents gender bias LIC scores for all models on ICE-B. The LIC score for the human-provided ground truth captions is 49.2. ClipCap performs best in terms of LIC. Both ClipCap and GIT captions are less biased than human-provided captions. This result suggests that certain language is being used for gender groups (i.e. terms such as `attractive', `young', `beautiful' only being used for images of female subjects). Once again the models (L-Verse, BITTERS) with more accurate captions semantically contain more gender bias. LIC scores in this range (40-50) have been observed for other captioning models \cite{hirota2022quantifying}, suggesting that most of image captioning approaches show comparable levels of gender bias.

\begin{table}[t]
\centering
\footnotesize
\addtolength{\tabcolsep}{-2pt}
\begin{threeparttable}[t]
\begin{tabular}{lcc}
\toprule
Metric            &  BITTERS     & \\ 
\midrule
Normalized Keyword Overlap (\%) $\uparrow$           &   35.7 &\\
CLIP Cosine Similarity (\%) $\uparrow$           &  37.6 &\\
\midrule
\end{tabular}

\end{threeparttable}
\caption
{
Keyword extraction performance on ICE-A.
}
\label{table:keyword_eval_set_a}
\vspace{-4mm}
\end{table}

\subsection{Zero-Shot Keyword Extraction}
Table \ref{table:keyword_eval_set_a} presents keyword extraction results on ICE-A. As described in Section \ref{sampling}, we currently sample 48 tokens from each image. Our BITTERS extracts a median of 15 keywords per image while the ground truth median is 41. Despite the lack of overlap with the ground truth, the model extracted keywords are observed to be highly relevant and successfully describe the contents of the image. We provide examples of extracted keywords along with generated captions in our supplementary material.

\section{Conclusion}
\label{sec:conclusion}
In this paper, we present a novel model architecture and a set of evaluation data and metrics for zero-shot image captioning. As we bring the bidirectional image-text training strategy \cite{kim2022verse} to large-scale, we find that a well-curated dataset of image-text pairs enables zero-shot image captioning. BITTERS is parameter-efficient architecture specially designed for zero-shot image captioning and keyword extraction. 
The proposed WaveVAE uses 2D DWT \cite{antonini1992dwt} for cross-level feature augmentation \cite{kim2022verse}. Our WaveVAE shows improved image reconstruction performance and adaptability to large-scale training compared to AugVAE. \cite{kim2022verse}. We further assess the accuracy and bias of zero-shot generated captions with BITTERS in both a quantitative and qualitative manner.
\section{Discussion} 
\label{sec:discussion}
\paragraph{Limitation}
Future work in this area will need to address the trade-off between societal bias and semantic accuracy. This can be accomplished through training set audits to mitigate bias and improve representation. The development of improved bias detection and mitigation techniques for caption generation will also be required. We also observe a trade-off between the number of keywords and the quality of the individual keyword. Further enhancing the keyword extraction capabilities of this model will enable robust image tagging for large-scale retrieval applications. Even when trained in a bidirectional manner, our model lacks text-to-image generation capability compared to other state-of-the-art models. We provide further discussion on \textit{zero-shot} and \textit{large-scale bidirectional training} in our supplementary material.

\paragraph{Broader Impact}
Our model can cover a variety of images with its zero-shot capability to help visually-impaired people. Although our training set (TIP100M) does not contain any toxic language, societal bias in captions generated with BITTERS should be properly mitigated before public use.

{\small
\bibliographystyle{ieee_fullname}
\bibliography{egbib}

\begin{thebibliography}{10}\itemsep=-1pt

\bibitem{anderson2016spice}
Peter Anderson, Basura Fernando, Mark Johnson, and Stephen Gould.
\newblock Spice: Semantic propositional image caption evaluation.
\newblock In {\em Proceedings of the European Conference on Computer Vision},
  2016.

\bibitem{antonini1992dwt}
M. Antonini, M. Barlaud, P. Mathieu, and I. Daubechies.
\newblock Image coding using wavelet transform.
\newblock {\em IEEE Transactions on Image Processing}, 1992.

\bibitem{banerjee2005meteor}
Satanjeev Banerjee and Alon Lavie.
\newblock Meteor: An automatic metric for mt evaluation with improved
  correlation with human judgments.
\newblock In {\em Proceedings of the acl workshop on intrinsic and extrinsic
  evaluation measures for machine translation and/or summarization}, 2005.

\bibitem{Brown2020LanguageMA}
Tom Brown, Benjamin Mann, Nick Ryder, Melanie Subbiah, Jared~D Kaplan, Prafulla
  Dhariwal, Arvind Neelakantan, Pranav Shyam, Girish Sastry, Amanda Askell,
  Sandhini Agarwal, Ariel Herbert-Voss, Gretchen Krueger, Tom Henighan, Rewon
  Child, Aditya Ramesh, Daniel Ziegler, Jeffrey Wu, Clemens Winter, Chris
  Hesse, Mark Chen, Eric Sigler, Mateusz Litwin, Scott Gray, Benjamin Chess,
  Jack Clark, Christopher Berner, Sam McCandlish, Alec Radford, Ilya Sutskever,
  and Dario Amodei.
\newblock Language models are few-shot learners.
\newblock In {\em Advances in Neural Information Processing Systems}, 2020.

\bibitem{changpinyo2021cc12m}
Soravit Changpinyo, Piyush Sharma, Nan Ding, and Radu Soricut.
\newblock {Conceptual 12M}: Pushing web-scale image-text pre-training to
  recognize long-tail visual concepts.
\newblock In {\em Proceedings of the IEEE/CVF Conference on Computer Vision and
  Pattern Recognition}, 2021.

\bibitem{cho2021vlt5}
Jaemin Cho, Jie Lei, Hao Tan, and Mohit Bansal.
\newblock Unifying vision-and-language tasks via text generation.
\newblock In {\em Proceedings of the International Conference on Machine
  Learning}, 2021.

\bibitem{imagenet_cvpr09}
J. Deng, W. Dong, R. Socher, L.-J. Li, K. Li, and L. Fei-Fei.
\newblock {ImageNet: A Large-Scale Hierarchical Image Database}.
\newblock In {\em Proceedings of the IEEE/CVF Conference on Computer Vision and
  Pattern Recognition}, 2009.

\bibitem{desai2021redcaps}
Karan Desai, Gaurav Kaul, Zubin Aysola, and Justin Johnson.
\newblock {RedCaps: Web-curated image-text data created by the people, for the
  people}.
\newblock In {\em NeurIPS Datasets and Benchmarks}, 2021.

\bibitem{ding2021cogview}
Ming Ding, Zhuoyi Yang, Wenyi Hong, Wendi Zheng, Chang Zhou, Da Yin, Junyang
  Lin, Xu Zou, Zhou Shao, Hongxia Yang, and Jie Tang.
\newblock Cogview: Mastering text-to-image generation via transformers.
\newblock In {\em Advances in Neural Information Processing Systems}, 2021.

\bibitem{ding2022cogview2}
Ming Ding, Wendi Zheng, Wenyi Hong, and Jie Tang.
\newblock Cogview2: Faster and better text-to-image generation via hierarchical
  transformers, 2022.

\bibitem{esser2021taming}
Patrick Esser, Robin Rombach, and Bjorn Ommer.
\newblock Taming transformers for high-resolution image synthesis.
\newblock In {\em Proceedings of the IEEE/CVF Conference on Computer Vision and
  Pattern Recognition}, 2021.

\bibitem{fan2018topk}
Angela Fan, Mike Lewis, and Yann Dauphin.
\newblock Hierarchical neural story generation.
\newblock In {\em Proceedings of the 56th Annual Meeting of the Association for
  Computational Linguistics (Volume 1: Long Papers)}, 2018.

\bibitem{gafni2022makeasense}
Oran Gafni, Adam Polyak, Oron Ashual, Shelly Sheynin, Devi Parikh, and Yaniv
  Taigman.
\newblock Make-a-scene: Scene-based text-to-image generation with human priors,
  2022.

\bibitem{NIPS2014_5423}
Ian Goodfellow, Jean Pouget-Abadie, Mehdi Mirza, Bing Xu, David Warde-Farley,
  Sherjil Ozair, Aaron Courville, and Yoshua Bengio.
\newblock Generative adversarial nets.
\newblock In {\em Advances in Neural Information Processing Systems}, 2014.

\bibitem{he2015prelu}
Kaiming He, Xiangyu Zhang, Shaoqing Ren, and Jian Sun.
\newblock Delving deep into rectifiers: Surpassing human-level performance on
  imagenet classification.
\newblock In {\em Proceedings of International Conference on Computer Vision},
  2015.

\bibitem{hendricks2018women}
Lisa~Anne Hendricks, Kaylee Burns, Kate Saenko, Trevor Darrell, and Anna
  Rohrbach.
\newblock Women also snowboard: Overcoming bias in captioning models.
\newblock In {\em Proceedings of the European Conference on Computer Vision},
  2018.

\bibitem{hessel-etal-2021-clipscore}
Jack Hessel, Ari Holtzman, Maxwell Forbes, Ronan Le~Bras, and Yejin Choi.
\newblock {CLIPS}core: A reference-free evaluation metric for image captioning.
\newblock In {\em Proceedings of the 2021 Conference on Empirical Methods in
  Natural Language Processing}, 2021.

\bibitem{Heusel2017GANsTB}
Martin Heusel, Hubert Ramsauer, Thomas Unterthiner, Bernhard Nessler, and Sepp
  Hochreiter.
\newblock Gans trained by a two time-scale update rule converge to a local nash
  equilibrium.
\newblock In {\em Advances in Neural Information Processing Systems}, 2017.

\bibitem{hirota2022quantifying}
Yusuke Hirota, Yuta Nakashima, and Noa Garcia.
\newblock Quantifying societal bias amplification in image captioning.
\newblock In {\em Proceedings of the IEEE/CVF Conference on Computer Vision and
  Pattern Recognition}, 2022.

\bibitem{ho2020ddpm}
Jonathan Ho, Ajay Jain, and Pieter Abbeel.
\newblock Denoising diffusion probabilistic models.
\newblock In {\em Advances in Neural Information Processing Systems}, 2020.

\bibitem{hoffmann2022chinchilla}
Jordan Hoffmann, Sebastian Borgeaud, Arthur Mensch, Elena Buchatskaya, Trevor
  Cai, Eliza Rutherford, Diego de~Las Casas, Lisa~Anne Hendricks, Johannes
  Welbl, Aidan Clark, Tom Hennigan, Eric Noland, Katie Millican, George van~den
  Driessche, Bogdan Damoc, Aurelia Guy, Simon Osindero, Karen Simonyan, Erich
  Elsen, Jack~W. Rae, Oriol Vinyals, and Laurent Sifre.
\newblock Training compute-optimal large language models, 2022.

\bibitem{Holtzman2020topp}
Ari Holtzman, Jan Buys, Li Du, Maxwell Forbes, and Yejin Choi.
\newblock The curious case of neural text degeneration.
\newblock In {\em Proceedings of the International Conference on Learning
  Representations}, 2020.

\bibitem{houlsby2019adapter}
Neil Houlsby, Andrei Giurgiu, Stanislaw Jastrzebski, Bruna Morrone, Quentin de
  Laroussilhe, Andrea Gesmundo, Mona Attariyan, and Sylvain Gelly.
\newblock Parameter-efficient transfer learning for {NLP}.
\newblock In {\em Proceedings of the International Conference on Machine
  Learning}, 2019.

\bibitem{hu2021lora}
Edward Hu, Yelong Shen, Phil Wallis, Zeyuan Allen-Zhu, Yuanzhi Li, Lu Wang, and
  Weizhu Chen.
\newblock Lora: Low-rank adaptation of large language models.
\newblock In {\em Proceedings of the International Conference on Learning
  Representations}, 2022.

\bibitem{hu2022lemon}
Xiaowei Hu, Zhe Gan, Jianfeng Wang, Zhengyuan Yang, Zicheng Liu, Yumao Lu, and
  Lijuan Wang.
\newblock Scaling up vision-language pre-training for image captioning.
\newblock In {\em Proceedings of the IEEE/CVF Conference on Computer Vision and
  Pattern Recognition}, 2022.

\bibitem{hutto2014vader}
Clayton Hutto and Eric Gilbert.
\newblock Vader: A parsimonious rule-based model for sentiment analysis of
  social media text.
\newblock In {\em Proceedings of the international AAAI conference on web and
  social media}, 2014.

\bibitem{kim2022verse}
Taehoon Kim, Gwangmo Song, Sihaeng Lee, Sangyun Kim, Yewon Seo, Soonyoung Lee,
  Seung~Hwan Kim, Honglak Lee, and Kyunghoon Bae.
\newblock L-verse: Bidirectional generation between image and text.
\newblock In {\em Proceedings of the IEEE/CVF Conference on Computer Vision and
  Pattern Recognition}, 2022.

\bibitem{kingma2014autoencoding}
Diederik~P Kingma and Max Welling.
\newblock Auto-encoding variational bayes.
\newblock In {\em Proceedings of the International Conference on Learning
  Representations}, 2014.

\bibitem{li2023decap}
Wei Li, Linchao Zhu, Longyin Wen, and Yi Yang.
\newblock Decap: Decoding {CLIP} latents for zero-shot captioning via text-only
  training.
\newblock In {\em International Conference on Learning Representations}, 2023.

\bibitem{li2020oscar}
Xiujun Li, Xi Yin, Chunyuan Li, Pengchuan Zhang, Xiaowei Hu, Lei Zhang, Lijuan
  Wang, Houdong Hu, Li Dong, Furu Wei, Yejin Choi, and Jianfeng Gao.
\newblock Oscar: Object-semantics aligned pre-training for vision-language
  tasks.
\newblock In {\em Proceedings of the European Conference on Computer Vision},
  2020.

\bibitem{lin2004rouge}
Chin-Yew Lin.
\newblock Rouge: A package for automatic evaluation of summaries.
\newblock In {\em Text summarization branches out}, 2004.

\bibitem{lin2015microsoft}
Tsung-Yi Lin, Michael Maire, Serge Belongie, Lubomir Bourdev, Ross Girshick,
  James Hays, Pietro Perona, Deva Ramanan, C.~Lawrence Zitnick, and Piotr
  Dollár.
\newblock Microsoft coco: Common objects in context.
\newblock In {\em Proceedings of the European Conference on Computer Vision},
  2014.

\bibitem{lin-etal-2020-exploring}
Zhaojiang Lin, Andrea Madotto, and Pascale Fung.
\newblock Exploring versatile generative language model via parameter-efficient
  transfer learning.
\newblock In {\em Findings of the Association for Computational Linguistics:
  EMNLP}, 2020.

\bibitem{liu2021ptuningv2}
Xiao Liu, Kaixuan Ji, Yicheng Fu, Zhengxiao Du, Zhilin Yang, and Jie Tang.
\newblock P-tuning v2: Prompt tuning can be comparable to fine-tuning
  universally across scales and tasks, 2021.

\bibitem{liu2021ptuning}
Xiao Liu, Yanan Zheng, Zhengxiao Du, Ming Ding, Yujie Qian, Zhilin Yang, and
  Jie Tang.
\newblock Gpt understands, too, 2021.

\bibitem{DBLP:journals/corr/abs-1711-05101}
Ilya Loshchilov and Frank Hutter.
\newblock Fixing weight decay regularization in adam.
\newblock In {\em Proceedings of the International Conference on Learning
  Representations}, 2018.

\bibitem{miyato2018spectral}
Takeru Miyato, Toshiki Kataoka, Masanori Koyama, and Yuichi Yoshida.
\newblock Spectral normalization for generative adversarial networks.
\newblock In {\em Proceedings of International Conference on Learning
  Representations}, 2018.

\bibitem{mokady2021clipcap}
Ron Mokady, Amir Hertz, and Amit~H Bermano.
\newblock Clipcap: Clip prefix for image captioning, 2021.

\bibitem{nichol2021glide}
Alex Nichol, Prafulla Dhariwal, Aditya Ramesh, Pranav Shyam, Pamela Mishkin,
  Bob McGrew, Ilya Sutskever, and Mark Chen.
\newblock Glide: Towards photorealistic image generation and editing with
  text-guided diffusion models, 2021.

\bibitem{papineni2002bleu}
Kishore Papineni, Salim Roukos, Todd Ward, and Wei-Jing Zhu.
\newblock Bleu: a method for automatic evaluation of machine translation.
\newblock In {\em Proceedings of the 40th annual meeting of the Association for
  Computational Linguistics}, 2002.

\bibitem{radford2021learning}
Alec Radford, Jong~Wook Kim, Chris Hallacy, Aditya Ramesh, Gabriel Goh,
  Sandhini Agarwal, Girish Sastry, Amanda Askell, Pamela Mishkin, Jack Clark,
  Gretchen Krueger, and Ilya Sutskever.
\newblock Learning transferable visual models from natural language
  supervision.
\newblock In {\em Proceedings of the International Conference on Machine
  Learning}, 2021.

\bibitem{radford2019language}
Alec Radford, Jeff Wu, Rewon Child, David Luan, Dario Amodei, and Ilya
  Sutskever.
\newblock Language models are unsupervised multitask learners, 2019.

\bibitem{ramesh2022dalle2}
Aditya Ramesh, Prafulla Dhariwal, Alex Nichol, Casey Chu, and Mark Chen.
\newblock Hierarchical text-conditional image generation with clip latents,
  2022.

\bibitem{ramesh2021zeroshot}
Aditya Ramesh, Mikhail Pavlov, Gabriel Goh, Scott Gray, Chelsea Voss, Alec
  Radford, Mark Chen, and Ilya Sutskever.
\newblock Zero-shot text-to-image generation.
\newblock In {\em Proceedings of the International Conference on Machine
  Learning}, 2021.

\bibitem{razavi2019generating}
Ali Razavi, Aaron van~den Oord, and Oriol Vinyals.
\newblock Generating diverse high-fidelity images with vq-vae-2.
\newblock In {\em Advances in Neural Information Processing Systems}, 2019.

\bibitem{rombach2021stablediffusion}
Robin Rombach, Andreas Blattmann, Dominik Lorenz, Patrick Esser, and Björn
  Ommer.
\newblock High-resolution image synthesis with latent diffusion models, 2021.

\bibitem{ho2022imagen}
Chitwan Saharia, William Chan, Saurabh Saxena, Lala Li, Jay Whang, Emily
  Denton, Seyed Kamyar~Seyed Ghasemipour, Burcu~Karagol Ayan, S.~Sara Mahdavi,
  Rapha~Gontijo Lopes, Tim Salimans, Jonathan Ho, David~J Fleet, and Mohammad
  Norouzi.
\newblock Photorealistic text-to-image diffusion models with deep language
  understanding, 2022.

\bibitem{schuhmann2021laion400m}
Christoph Schuhmann, Richard Vencu, Romain Beaumont, Robert Kaczmarczyk,
  Clayton Mullis, Aarush Katta, Theo Coombes, Jenia Jitsev, and Aran
  Komatsuzaki.
\newblock Laion-400m: Open dataset of clip-filtered 400 million image-text
  pairs, 2021.

\bibitem{sennrich2016neural}
Rico Sennrich, Barry Haddow, and Alexandra Birch.
\newblock Neural machine translation of rare words with subword units.
\newblock In {\em Proceedings of the Annual Meeting of the Association for
  Computational Linguistics}, 2016.

\bibitem{sharma2018conceptual}
Piyush Sharma, Nan Ding, Sebastian Goodman, and Radu Soricut.
\newblock Conceptual captions: A cleaned, hypernymed, image alt-text dataset
  for automatic image captioning.
\newblock In {\em Proceedings of the Annual Meeting of the Association for
  Computational Linguistics}, 2018.

\bibitem{shi2016pixelshuffle}
Wenzhe Shi, Jose Caballero, Ferenc Huszár, Johannes Totz, Andrew~P. Aitken,
  Rob Bishop, Daniel Rueckert, and Zehan Wang.
\newblock Real-time single image and video super-resolution using an efficient
  sub-pixel convolutional neural network, 2016.

\bibitem{smith2017onecycle}
Leslie~N. Smith and Nicholay Topin.
\newblock Super-convergence: Very fast training of residual networks using
  large learning rates, 2017.

\bibitem{tao2021dfgan}
Ming Tao, Hao Tang, Songsong Wu, Nicu Sebe, Xiao-Yuan Jing, Fei Wu, and Bingkun
  Bao.
\newblock Df-gan: Deep fusion generative adversarial networks for text-to-image
  synthesis, 2021.

\bibitem{tewel2022zerocap}
Yoad Tewel, Yoav Shalev, Idan Schwartz, and Lior Wolf.
\newblock Zerocap: Zero-shot image-to-text generation for visual-semantic
  arithmetic.
\newblock In {\em Proceedings of the IEEE/CVF Conference on Computer Vision and
  Pattern Recognition (CVPR)}, pages 17918--17928, June 2022.

\bibitem{thomee2016yfcc}
Bart Thomee, David~A. Shamma, Gerald Friedland, Benjamin Elizalde, Karl Ni,
  Douglas Poland, Damian Borth, and Li-Jia Li.
\newblock {YFCC}100m.
\newblock {\em Communications of the {ACM}}, 2016.

\bibitem{oord2018neural}
Aaron van~den Oord, Oriol Vinyals, and Koray Kavukcuoglu.
\newblock Neural discrete representation learning.
\newblock In {\em Advances in Neural Information Processing Systems}, 2017.

\bibitem{vedantam2015cider}
Ramakrishna Vedantam, C.~Lawrence Zitnick, and Devi Parikh.
\newblock Cider: Consensus-based image description evaluation.
\newblock In {\em Proceedings of the IEEE Conference on Computer Vision and
  Pattern Recognition}, 2015.

\bibitem{wang2022git}
Jianfeng Wang, Zhengyuan Yang, Xiaowei Hu, Linjie Li, Kevin Lin, Zhe Gan,
  Zicheng Liu, Ce Liu, and Lijuan Wang.
\newblock Git: A generative image-to-text transformer for vision and language,
  2022.

\bibitem{wang2022ofa}
Peng Wang, An Yang, Rui Men, Junyang Lin, Shuai Bai, Zhikang Li, Jianxin Ma,
  Chang Zhou, Jingren Zhou, and Hongxia Yang.
\newblock {OFA}: Unifying architectures, tasks, and modalities through a simple
  sequence-to-sequence learning framework.
\newblock In {\em Proceedings of the International Conference on Machine
  Learning}, 2022.

\bibitem{wang2021realesrgan}
Xintao Wang, Liangbin Xie, Chao Dong, and Ying Shan.
\newblock Real-esrgan: Training real-world blind super-resolution with pure
  synthetic data.
\newblock In {\em International Conference on Computer Vision Workshops
  (ICCVW)}, 2021.

\bibitem{xu2017attngan}
Tao Xu, Pengchuan Zhang, Qiuyuan Huang, Han Zhang, Zhe Gan, Xiaolei Huang, and
  Xiaodong He.
\newblock Attngan: Fine-grained text to image generation with attentional
  generative adversarial networks.
\newblock In {\em Proceedings of the IEEE/CVF Conference on Computer Vision and
  Pattern Recognition}, 2018.

\bibitem{yu2022vectorquantized}
Jiahui Yu, Xin Li, Jing~Yu Koh, Han Zhang, Ruoming Pang, James Qin, Alexander
  Ku, Yuanzhong Xu, Jason Baldridge, and Yonghui Wu.
\newblock Vector-quantized image modeling with improved {VQGAN}.
\newblock In {\em Proceedings of the International Conference on Learning
  Representations}, 2022.

\bibitem{yu2022parti}
Jiahui Yu, Yuanzhong Xu, Jing~Yu Koh, Thang Luong, Gunjan Baid, Zirui Wang,
  Vijay Vasudevan, Alexander Ku, Yinfei Yang, Burcu~Karagol Ayan, Ben
  Hutchinson, Wei Han, Zarana Parekh, Xin Li, Han Zhang, Jason Baldridge, and
  Yonghui Wu.
\newblock Scaling autoregressive models for content-rich text-to-image
  generation, 2022.

\bibitem{zhang2021xmc}
Han Zhang, Jing~Yu Koh, Jason Baldridge, Honglak Lee, and Yinfei Yang.
\newblock Cross-modal contrastive learning for text-to-image generation.
\newblock In {\em Proceedings of the IEEE/CVF Conference on Computer Vision and
  Pattern Recognition}, 2021.

\bibitem{zhang2021vinvl}
Pengchuan Zhang, Xiujun Li, Xiaowei Hu, Jianwei Yang, Lei Zhang, Lijuan Wang,
  Yejin Choi, and Jianfeng Gao.
\newblock Vinvl: Revisiting visual representations in vision-language models.
\newblock In {\em Proceedings of the IEEE/CVF Conference on Computer Vision and
  Pattern Recognition}, 2021.

\bibitem{zhang2018unreasonable}
Richard Zhang, Phillip Isola, Alexei~A. Efros, Eli Shechtman, and Oliver Wang.
\newblock The unreasonable effectiveness of deep features as a perceptual
  metric.
\newblock In {\em Proceedings of the IEEE/CVF Conference on Computer Vision and
  Pattern Recognition}, 2018.

\bibitem{zhao2021understanding}
Dora Zhao, Angelina Wang, and Olga Russakovsky.
\newblock Understanding and evaluating racial biases in image captioning.
\newblock In {\em Proceedings of the IEEE/CVF International Conference on
  Computer Vision}, 2021.

\bibitem{zhu2019dmgan}
Minfeng Zhu, Pingbo Pan, Wei Chen, and Yi Yang.
\newblock Dm-gan: Dynamic memory generative adversarial networks for
  text-to-image synthesis.
\newblock In {\em Proceedings of the IEEE/CVF Conference on Computer Vision and
  Pattern Recognition}, 2019.

\end{thebibliography}
}

\clearpage
\appendix
\section{Related Work}
\subsection{Parameter-Efficient Finetuning}
Finetuning large language models (LLMs) from scratch requires huge computing resources.  To efficiently finetune LLMs, several parameter-efficient finetuning approaches have been proposed.  Prompt tuning (p-tuning) \cite{liu2021ptuning, liu2021ptuningv2} adds a small encoder to an LLM to generate an appropriate input prompt for each downstream task. Adapter-based approaches \cite{houlsby2019adapter, hu2021lora, lin-etal-2020-exploring} add a residual, bottleneck-style adapter to each layer of an LLM and only update adapters during finetuning. Unlike from-scratch finetuning, updating a smaller portion of an LLM can greatly reduce memory usage and computing time.

\subsection{Bidirectional Auto-Regressive Transformer}
L-Verse \cite{kim2022verse} is first proposed as a bidirectional model that can generate image from text and vice versa. Along with other \texttt{DALL-E} \cite{ramesh2021zeroshot} variants \cite{ding2021cogview, ding2022cogview2, gafni2022makeasense,yu2022parti}, encodes an image into a sequence of tokens to utilize the scalability of auto-regressive transformer architecture \cite{radford2021learning}. Bidirectional Auto-Regressive Transformer (BiART) \cite{kim2022verse} uses segment embedding to distinguish between image (or text) as a
conditional reference and a generation target. Unlike other models \cite{ramesh2021zeroshot, ding2021cogview}, BiART doesn't require extra optimization techniques to enable \texttt{FP16(O2)} automatic-mixed-precision (AMP) training. 

\section{Method}

\subsection{WaveVAE}

\paragraph{Training}
For both stages, we train WaveVAE with AdamW \cite{DBLP:journals/corr/abs-1711-05101} optimizer with $\beta_1 = 0.9$, $\beta_2 = 0.999$, $\epsilon = 10e-8$. We only apply weight decay in Stage 1 with weight decay multiplier of $1e-5$. We use learning rate $3.6e-5$ and apply linear learning rate warm-up for the first 1\% of iterations and then decay the learning rate to $3.6e-6$ using cosine learning rate decay \cite{smith2017onecycle}.  We also resize each image to $256 \times 256 \times 3$ and apply random crop with 0.75 crop ratio.
\paragraph{Stage 1} As depicted in Figure \ref{wavevae_figure}, we first pretrain pairs of encoders and decoders with 2D DWT (\texttt{Haar}) approximations of an input image in different resolutions. $L_1$ losses between the original and reconstructed image of each pair are summed and used as a loss term to update the model. We train the model for 3 epochs with a batch size of 480.

\paragraph{Stage 2} After architecture modification and a small calibration, we further train WaveVAE with a weighted sum of $L_1$, LPIPS \cite{zhang2018unreasonable}, and adversarial \cite{wang2021realesrgan, esser2021taming} losses. Unlike VQGAN \cite{esser2021taming}, we use a U-Net discriminator \cite{wang2021realesrgan} with spectral normalization \cite{miyato2018spectral}. Replacing the discriminator and multiplying the adversarial loss by $1.0e-3$ allow stable training on both ImageNet1K \cite{imagenet_cvpr09} and TIP100M without hyperparameter changes. We train the model for 10 epochs with batch size 3840.

\subsection{BiART}
\paragraph{Training}
With the encoder part of WaveVAE, we train BiART on the 100 million image-caption pairs of TIP100M following the bidirectional training process proposed in \cite{kim2022verse}. We train BiART in \texttt{FP16(O2)} automatic-mixed-precision (AMP). Unlike L-Verse \cite{kim2022verse}, there is no need to perform inference of WaveVAE in \texttt{FP32} full-precision to prevent the underflow. We use AdamW \cite{DBLP:journals/corr/abs-1711-05101} optimizer with $\beta_1 = 0.9$, $\beta_2 = 0.95$, $\epsilon = 1e-8$, weight decay multiplier $1e-2$, and learning rate $1.5e-4$. We don't apply weight decay to embedding parameters. We train our model for 2 epochs in total with batch size 1280. We apply linear learning rate warm-up for first 1\% of iterations and then decay the learning rate to $1.5e-5$ using cosine learning rate decay \cite{smith2017onecycle}. We directly use the pretrained BITTERS model for zero-shot image captioning.

\section{Dataset}
\subsection{Training (TIP100M)}

\paragraph {Details and Publicity} Each image included in TIP100M is \textbf{100\% licensed and was approved via Shutterstock's\footnote{\href{https://www.shutterstock.com}{www.shutterstock.com}} human review system, which controls for image quality and legal compliance}. The dataset is random sampled from Shutterstock's image catalog to capture an incredibly broad set of visual concepts as mentioned in Section \ref{sec:dataset_tip}. The catalog with watermarked images and corresponding metadata is open to public. We do not own any of original images in TIP100M and hence cannot legally provide them to public \textit{as-is}. \textbf{We will instead provide official links\footnote{\href{https://www.shutterstock.com/image-photo/red-apple-isolated-on-white-background-1727544364}{Example: red apple isolated on white background.}} to all images (watermarked) and corresponding metadata upon acceptance.}

\paragraph {Importance}
While previous works \cite{ramesh2021zeroshot,ding2021cogview,ding2022cogview2, ramesh2022dalle2, nichol2021glide, ho2022imagen,yu2022parti} focus on the quantity of training data, we put more emphasis on the quality. As we mentioned in Sections \ref{sec:intro} and \ref{sec:related_work}, we believe the quality of each caption is the key to zero-shot image captioning. Along with high-quality ground-truth captions, we also provide a list of keywords for each image to further promote research on different zero-shot vision-language tasks including zero-shot keyword extraction, image tagging, image retrieval, and keyword-to-image generation.

\subsection{Evaluation (ICE-A and B)}

\paragraph {Details and Publicity} Unlike TIP100M, images in ICE-A and B are licensed under \texttt{CC BY-NC-ND 4.0}, \textbf{which allows anyone to copy and redistribute the material in any medium or format}. We carefully selected each images by criteria in Section \ref{sec:dataset_ice}.  \textbf{We already opened download links for ICE-A and B to the public}. Please understand that we can't include specific links in this version to keep anonymity. Furthermore, we are currently hosting a global challenge on zero-shot image captioning with ICE-A and B along with the evaluation server to promote future researches on zero-shot image captioning. The result of the challenge will also be included in our final version.

\paragraph {Importance}
ICE-A and B includes a larger variety of visual concepts from many domains as well as various image types (photographs, illustrations, graphics). While large-scale training sets contain various types of images collected from the web, benchmark evaluation sets mainly contain real photographs. To evaluate a model's true captioning performance, ICE-A and B is essential. As far as we know, we are also the first to release an evaluation set for societal bias. Previous works \cite{hendricks2018women,zhao2021understanding,hirota2022quantifying} only propose metrics to evaluate bias in existing benchmark datasets.

\section{Metrics}
\subsection{Caption Accuracy}
\begin{figure}[!tb]
  \centering
\includegraphics[width=\linewidth]{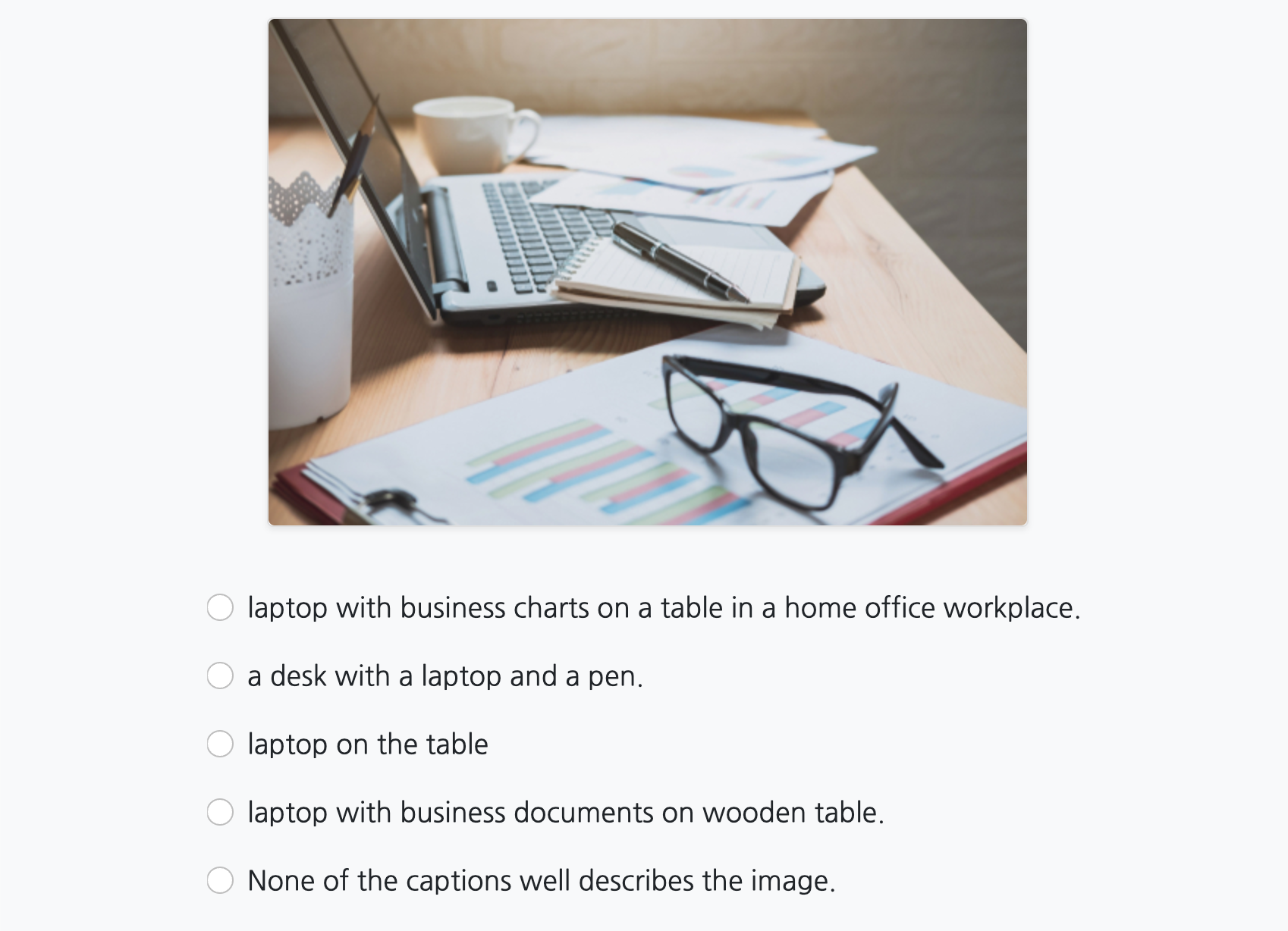}
\caption{Example interface for human evaluation.}
\label{eval_interface}

\end{figure}

Along with overall results on five metrics mentioned in Section \ref{sec:metric}, we also provide SPICE per image category (ICE-A) and ethnicity group (ICE-B) for detailed examination. For human evaluation, we use a web-based human evaluation
tool as shown in Figure \ref{eval_interface}.

\subsection{Bias Assessment}
Following three metrics are used to assess the societal bias of a zero-shot image captioning model. Details including gender terms and usage may differ from the original.
\begin{itemize}
    \item \textbf{Gender Error and Term Ratio}: Proposed for bias assessment in \cite{hendricks2018women}. Gender error is the rate of incorrect gender term (e.g. `man', `woman') usage in the set of generated captions. Gender term ratio is the ratio of female-terms to male-terms within the set of generated captions. High gender error suggests that a model is biased. This is potentially due to societal stereotypes in the training data (e.g. certain professions or clothing being associated with a given gender). The gender ratio should be as close as possible to the actual ratio of female-subjects to male-subjects in the evaluation set. Full lists of the gender terms used are provided below.
    \item \textbf{VADER Sentiment Score}: Proposed for caption bias assessment in \cite{zhao2021understanding}. The VADER language model \cite{hutto2014vader} is used to produce a compound sentiment score for a given image caption between -1.0 (very negative) and 1.0 (very positive). This score is influenced by the occurrence of sentiment-heavy terms such as `happy', `sad' or `angry'. The score is considered neutral if it lies between -0.05 and 0.05 \cite{zhao2021understanding}. In this paper, we compare the neutral sentiment rate of generated captions between gender groups and between ethnicity groups. Large differences in sentiment rate between gender (or ethnicity) groups is deemed to be undesirable and suggests that model is biased (i.e. emotive language only used for images of certain demographic groups).
    \item \textbf{Leakage for Image Captioning (LIC)}: Proposed in \cite{hirota2022quantifying}. Following the implementation of the original paper, we remove all gender terms from each caption (lists provided below) and train a language model (LSTM) which performs binary classification between genders. If the trained classifier can accurately predict the gender without these protected terms, bias is present in the caption (i.e. certain language used only for a certain group such as the word `attractive' only being used for women). The LIC score is then calculated as the gender classifier accuracy weighted by posterior probability.  Higher LIC indicates more biased captions.
    Hirota \etal \cite{hirota2022quantifying} use LIC score to evaluate bias amplification from training data to a captioning models output. As we do not train and test on the same samples, we slightly modify the usage of this method. We instead focus on measuring LIC to directly compare the bias in model-generated captions against the bias in human-labeled ground-truth captions.
\end{itemize}
The following gender terms are used for bias assessment: 
\begin{itemize}
\item \textbf{Male}: man, men, male, father, gentleman, gentlemen, boy, boys, uncle, husband, prince, waiter, son, he, his, him, himself, brother, brothers, guy, guys, emperor, emperors, dude, dudes, cowboy, businessman, policeman. 

\item \textbf{Female}: woman, women, female, lady, policewoman, ladies, mother, girl, girls, aunt, wife, actress, lesbian, princess, waitress, daughter, she, her, hers, herself, sister, sisters, queen, queens, pregnant, businesswoman, businesslady.
\end{itemize}

\subsection{Keyword Extraction}
\begin{itemize}
    \item \textbf{Normalized Keyword Overlap} The mean percentage (\%) of model extracted keywords found within the ground truth keywords for each image. 
    \item \textbf{CLIP Cosine Similarity} The mean percentage (\%) of model extracted keywords per image that have a text-image CLIP vector cosine similarity \cite{radford2021learning} exceeding a given threshold (0.23). This threshold was qualitatively determined by calculating a mean for the overall image-keywords pairs in ICE-A.
\end{itemize}

\section{Experiment}
\subsection{Text-To-Image Generation}
Table \ref{table:gen_fid} presents text-to-image generation performance of BITTERS. While text-to-image generation task is out of scope of this work, we provide the result to give an insight for future works on large-scale bidirectional training for zero-shot text-to-image generation. 

\paragraph{Sampling}
We modify the image sampling process proposed in \cite{kim2022verse} to delicately control the generated image. We sample 1024 image tokens with pretrained BITTERS model to generate an image for each text. For each token selection, we first select 10\% of logits with the highest probabilities (\textit{top-k} sampling) \cite{fan2018topk} and apply \textit{top-p} sampling \cite{Holtzman2020topp} with $p=0.95$. Since our model is bidirectionally trained, our model also supports classifier-free guidance for auto-regressive transformers used in \cite{gafni2022makeasense} without additional finetuning. We apply classifier-free guidance with the guidance scale $\alpha_c = 5$. We sample 64 images in total and calculate CLIPScore \cite{hessel-etal-2021-clipscore} to select a Top-1 image. 

\paragraph{Fréchet Inception Distance}
 We evaluate the text-to-image generation performance of BITTERS with Fréchet Inception Distance (FID) on a subset of 30,000 captions sampled from MS-COCO Captions validation set in Table \ref{table:gen_fid}. Following previous transformer-based models \cite{ramesh2021zeroshot, ding2021cogview, kim2022verse}, we compute FIDs after applying a Gaussian filter with varying radii to both original and generated images. Same with L-Verse \cite{kim2022verse}, our BITTERS shows decreasing FID with
increasing blur radius. Compared to L-Verse trained on Conceptual Captions \cite{sharma2018conceptual} (L-Verse-CC3M), BITTERS shows overall enhancement due to the increased number of training data. \textbf{Scaling up the number of parameters or training samples can be considered to improve our model for zero-shot text-to-image generation.} 
\begin{table}[!tb]
\centering
\footnotesize
\addtolength{\tabcolsep}{-2pt}
\begin{threeparttable}[t]
\begin{tabular}{lcccccc}
\toprule
Model               & FID-0         & FID-1 & FID-2 & FID-4 &  FID-8 & \\ 
\midrule
AttnGAN \cite{xu2017attngan}            & 35.2           & 44.0           & 72.0          & 108.0         & 100.0 & \\
DM-GAN \cite{zhu2019dmgan}             & 26.0  & 39.0           & 73.0          & 119.0         & 112.3 & \\
DF-GAN \cite{tao2021dfgan}             & 26.0  & 33.8  & 55.9          & 91.0          & 97.0  &\\
XMC-GAN \cite{zhang2021xmc}             & 9.33  & -           & -          & -          & -  & \\
\midrule
L-Verse-COCO \cite{kim2022verse}                               & 45.8           &  41.9         & 35.5 & 30.2 &  29.8 &\\
L-Verse-CC3M \cite{kim2022verse}           & 37.2          &  31.6          &  25.7         & 21.4        &  21.1 & \\
BITTERS            & 28.7          &  22.5          &  14.8         & 13.9        &  13.4 & \\
\midrule
DALL-E \cite{ramesh2021zeroshot}             & 27.5           & 28.0           & 45.5          & 83.5          & 85.0  & \\
CogView \cite{ding2021cogview}            & 27.1           & 19.4           & 13.9          & 19.4          & 23.6  & \\
GLIDE \cite{nichol2021glide}        & 12.24       & -           & -          & -          & -  & \\
Make-A-Sense \cite{gafni2022makeasense}        & 11.84        & -           & -          & -          & -  & \\
DALL-E 2 \cite{ramesh2022dalle2}      & 10.39   & -           & -          & -          & -  & \\
Cogview 2 \cite{ding2022cogview2}       & 27.5      & -           & -          & -          & -  & \\
Imagen \cite{ho2022imagen}             & 7.27          & -           & -          & -          & -  & \\
Parti \cite{yu2022parti}             & 7.23          & -           & -          & -          & -  & \\
\midrule
\end{tabular}
\begin{tablenotes}
  \item[*] \footnotesize{\textbf{FID-$k$}: FID of images blurred by radius $k$ Gaussian filter.}
\end{tablenotes}

\end{threeparttable}
\vspace*{1mm}
\caption
{
Fréchet Inception Distance (FID) on a subset of 30,000 captions sampled from MS-COCO Captions validation set.
}
\label{table:gen_fid}
\vspace*{-1mm}
\end{table}
\section{Discussion}

\paragraph{Zero-Shot}
We follow the definition of zero-shot proposed in Ramesh \etal \cite{ramesh2021zeroshot}, which is to \textbf{train a model with a large-scale dataset and use the model for evaluation on various datasets without additional finetuning} (\textit{cross-dataset}). This is different from the zero-shot concept mentioned in previous works, to finetune a model with MS-COCO Captions and evaluate on Flickr30k (\textit{cross-domain}). 

\paragraph{Large-Scale Bidirectional Training}
Our WaveVAE is designed for large-scale training, along with better performance and efficiency. As mentioned in Section \ref{sec:exp_recon}, WaveVAE shows high performance improvement with larger dataset, while AugVAE shows performance degradation. 

As shown in Table \ref{table:gen_fid}, our model doesn't show enough performance to compete with state-of-the-art zero-shot text-to-image generation models \cite{ding2022cogview2, ramesh2022dalle2, nichol2021glide, ho2022imagen,yu2022parti}. 
While text-to-image generation models focus on \textit{scale}, we rather focus on \textit{efficiency}. Compared to other transformer-based text-to-image generation models \cite{ding2021cogview, ding2022cogview2, gafni2022makeasense, yu2022parti} which require more than 3 billion parameters for zero-shot text-to-image generation, 650 million parameters are enough for zero-shot image captioning. \textbf{As text-to-image generation is out of scope of this work, we keep the number of parameters as small as possible}. 

\onecolumn
\section{Examples for Zero-Shot Image Captioning and Keyword Extraction}
\label{cap_keyword_ex}
\begin{figure*}[!hb]
\vspace*{-0.05in}
  \centering
\includegraphics[width=0.95\linewidth]{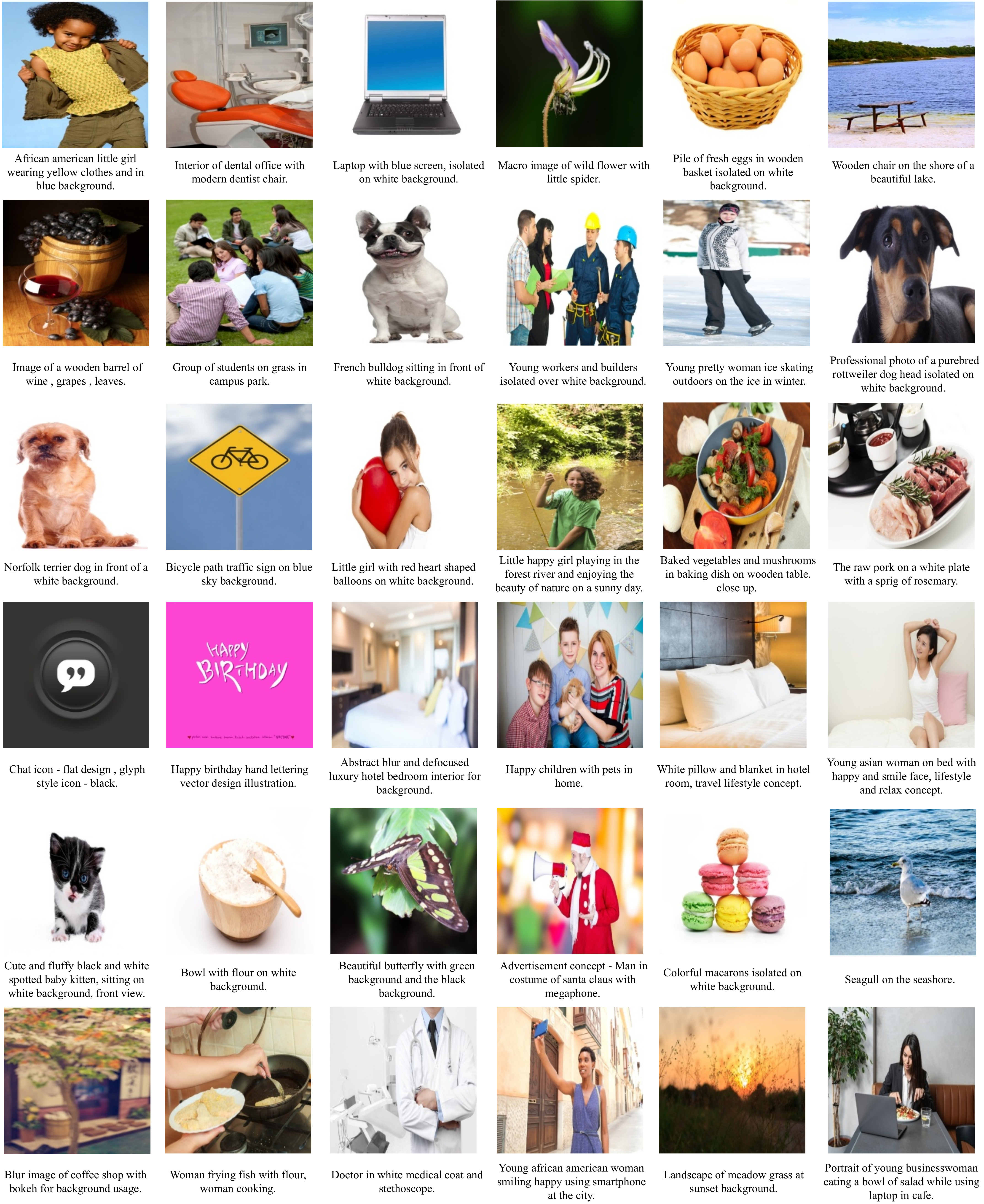}

\caption{
Examples of generated captions on ICE-A. Images are identical to images in Figure \ref{more_keywords}.  }
\label{more_captions}
\vspace{-4mm}
\end{figure*}

\begin{figure*}[!hb]
\vspace*{-0.05in}
  \centering
\includegraphics[width=0.95\linewidth]{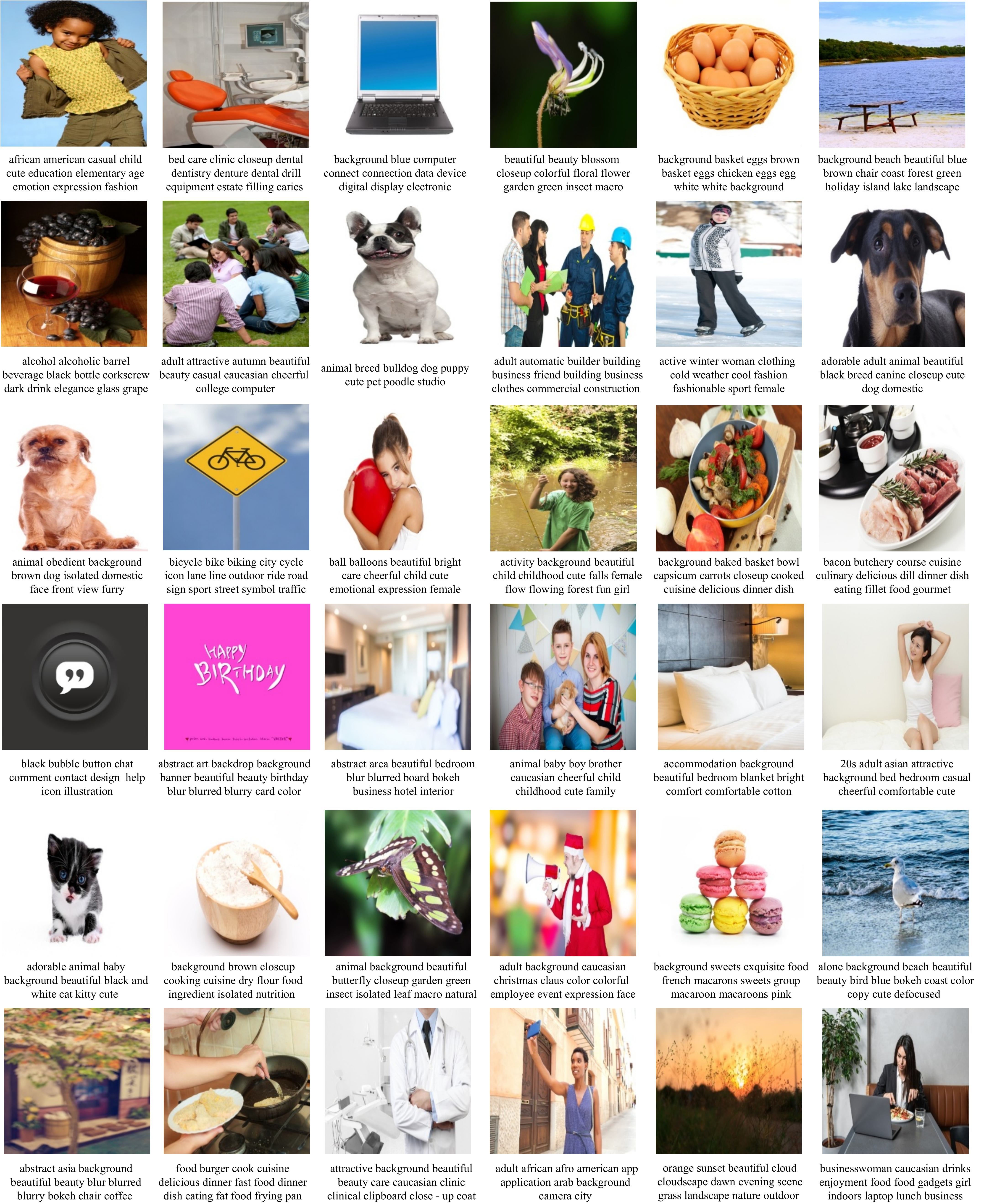}

\caption{
Examples of extracted keywords on ICE-A. Images are identical to images in Figure \ref{more_captions}. }
\label{more_keywords}
\vspace{-4mm}
\end{figure*}

We use the same images for Figures \ref{more_captions} and \ref{more_keywords} to show the relevance between a generated caption and generated list of keywords for a given image. 

\end{document}